\RequirePackage{fix-cm}
\documentclass{svjour3}                     % onecolumn (standard format)
\smartqed  % flush right qed marks, e.g. at end of proof
\usepackage{amsmath}
\usepackage{amsfonts}
\usepackage{graphicx}
\usepackage{subfigure}
\usepackage{natbib}
\usepackage{url}
\usepackage{caption}
\usepackage{footnote}
\makesavenoteenv{tabular}
\makesavenoteenv{table}

\usepackage[graphicx]{realboxes}
\DeclareMathOperator*{\argminA}{arg\,min}
\begin{document}

\title{Transfer Learning for Clinical Time Series Analysis using Deep Neural Networks}


\author{Priyanka Gupta \and Pankaj Malhotra \and Jyoti Narwariya \and Lovekesh Vig \and Gautam Shroff  %\and
}


\institute{
              TCS Research, New Delhi, India\\
              \email{\{priyanka.g35,malhotra.pankaj,jyoti.narwariya,lovekesh.vig,gautam.shroff\}@tcs.com}           %  \\
}

\date{Received: date / Accepted: date}

\maketitle

\begin{abstract}
Deep neural networks have shown promising results for various clinical prediction tasks such as diagnosis, mortality prediction, predicting duration of stay in hospital, etc.
However, training deep networks – such as those based on Recurrent Neural Networks (RNNs) – requires large labeled data, significant hyper-parameter tuning effort and expertise, and high computational resources.
In this work, we investigate as to what extent can transfer learning address these issues when using deep RNNs to model multivariate clinical time series. 
We consider two scenarios for transfer learning using RNNs: 
i) \textit{domain-adaptation}, i.e., leveraging a deep RNN -- namely, \textit{TimeNet} -- pre-trained for feature extraction on time series from diverse domains, and adapting it for feature extraction and subsequent target tasks in healthcare domain, 
ii) \textit{task-adaptation}, i.e., pre-training a deep RNN -- namely, \textit{HealthNet} -- on diverse tasks in healthcare domain, and adapting it to new target tasks in the same domain.
We evaluate the above approaches on publicly available MIMIC-III benchmark dataset, and demonstrate that (a) computationally-efficient linear models trained using features extracted via pre-trained RNNs outperform or, in the worst case, perform as well as deep RNNs and statistical hand-crafted features based models trained specifically for target task; 
(b) models obtained by adapting pre-trained models for target tasks are significantly more robust to the size of labeled data compared to task-specific RNNs, while also being computationally efficient.
We, therefore, conclude that pre-trained deep models like TimeNet and HealthNet allow leveraging the advantages of deep learning for clinical time series analysis tasks, while also minimize dependence on hand-crafted features, deal robustly with scarce labeled training data scenarios without overfitting, as well as reduce dependence on expertise and resources required to train deep networks from scratch (e.g. neural network architecture selection and hyper-parameter tuning efforts).
\end{abstract}
\keywords{Transfer learning, RNN, EHR data, Clinical Time Series, Patient Phenotyping, In-hospital Mortality Prediction, TimeNet }

\section{Introduction\label{intro}}
Electronic health records (EHR) consisting of a patient's medical history can be leveraged for various clinical applications such as diagnosis, recommending medicine, etc. Traditional machine learning techniques often require careful domain-specific feature engineering before building the prediction models.
On the other hand, deep learning approaches enable end-to-end learning without the need of hand-crafted and domain-specific features, and have recently produced promising results for various clinical prediction tasks \citep{lipton2015learning,miotto2017deep,ravi2017deep}.
Given this, there has been a rapid growth in the applications of deep learning to various clinical prediction tasks from Electronic Health Records, e.g. Doctor AI \citep{choi2016doctor} for medical diagnosis, Deep Patient \citep{miotto2016deep} to predict future diseases in patients, DeepR \citep{nguyen2017mathtt}  to predict unplanned readmission after discharge, etc.
With various medical parameters being recorded over a period of time in EHR databases, Recurrent Neural Networks (RNNs) can be an effective way to model the sequential aspects of EHR data, e.g. diagnoses \citep{lipton2015learning,che2016recurrent,choi2016doctor}, mortality prediction and estimating length of stay \citep{harutyunyan2017multitask,purushotham2017benchmark,rajkomar2018scalable}.

However, RNNs require large labeled data for training like any other deep learning approach and are prone to overfitting when labeled training data is scarce, and often require careful and computationally-expensive hyper-parameter tuning effort.
Transfer learning \citep{pan2010survey,bengio2012deep} has been demonstrated to be useful to address some of these challenges.
It enables knowledge transfer from neural networks trained on a \textit{source} task (domain) with sufficient training instances to a related \textit{target} task (domain) with few training instances.
For example, training a deep network on diverse set of images can provide useful features for images from unseen domains \citep{simonyan2014very}.
Moreover, fine-tuning a pre-trained network for target task is often faster and easier than constructing and training a new network from scratch \citep{bengio2012deep,malhotra2017timenet}.

It has been shown that pre-trained networks can learn to extract a rich set of generic features that can then be applied to a wide range of other similar tasks \citep{malhotra2017timenet}.
Also, it has been argued that transferring weights even from distant tasks can be better than using random initial weights \citep{yosinski2014transferable}.
Transfer learning via fine-tuning parameters of pre-trained models for end tasks has been recently considered for medical applications as well, e.g. \citep{choi2016doctor,lee2017transfer}.
However, fine-tuning a large number of parameters with a small labeled dataset may still result in overfitting, and requires careful regularization (as we show in Section \ref{ssec:healthnet} through empirical evaluation). 

In this work, we propose two simple yet effective approaches to transfer the knowledge captured in pre-trained deep RNNs for new target tasks in healthcare domain. 
More specifically, we consider two scenarios: i) extract features from a pre-trained network and use them to build models for target task, ii) initialize deep network for target task using parameters of a pre-trained network and then fine-tune using labeled training data for target task.
%If the parameters to be tuned for target task can be reduced to a small number, then the pre-trained deep models can be leveraged in a better way \citep{keshari2018learning}.
The key contributions of this work are: 
\begin{itemize}
	\item We propose two approaches for transfer learning using deep RNNs for classification tasks such as patient phenotyping and mortality prediction given the multivariate time series corresponding to physiological parameters of patients. We show effective approaches to adapt: i) general-purpose time series feature extractor based on deep RNNs (TimeNet, \cite{malhotra2017timenet}, detailed in Section \ref{ssec:approach1}) while overcoming the need to train a deep neural network from scratch while still leveraging its advantages, ii) a deep RNN trained specifically for healthcare domain (HealthNet, detailed in Section \ref{ssec:approach2}) while requiring significantly lesser amount of labeled training data for target task and significantly small hyper-parameter tuning effort.
	\item Our proposed approaches allow to extract robust features from variable length multivariate time series by using pre-trained deep RNNs, thereby reducing dependence on expert domain-driven feature extraction. We demonstrate that simple linear classification models trained on time series features extracted from our pre-trained models yield significantly better results compared to models trained using carefully extracted statistical features or deep RNN models trained from scratch specifically for the target task.
	\item We show that carefully regularized fine-tuning of pre-trained RNNs leads to models that are significantly more robust to training data sizes, and yield models that are significantly better compared to task-specific deep as well as shallow classification models trained from scratch, especially when training data is small.
\end{itemize}

Through empirical evaluation on patient phenotyping and mortality predictions tasks on MIMIC-III benchmark dataset \citep{johnson2016mimic} (as described in Section \ref{ssec:timenet} and \ref{ssec:healthnet}), we demonstrate that our transfer learning approaches yield data- and compute-efficient classification models that require little training effort while yielding performance comparable to models with hand-crafted features or carefully trained domain-specific deep networks benchmarked in \citep{harutyunyan2017multitask,song2017attend}.
%Further, adapting HealthNet models to a new target task is robust to training data sizes and yield models that are significantly better compared to task-specific deep as well as shallow classification models trained from scratch when training data is small.
\\

The rest of the paper is organized as follows: In Section \ref{sec:rw} we present some related work, and describe details of TimeNet in Section \ref{sec:bg}. We provide an overview of the proposed approaches in Section \ref{ssec:approach_overview},
and provide their details in Section \ref{ssec:approach1} and Section \ref{ssec:approach2}. 
We provide cohort selection and other details of dataset considered in Section \ref{ssec:data_desc}. We provide experimental details and observations made in Section \ref{ssec:timenet} and Section \ref{ssec:healthnet} respectively, and finally conclude in Section \ref{ssec:conclusion}.

\section{Related Work\label{sec:rw}}
\textit{Transfer Learning via Feature Extraction}
TimeNet-based features have been shown to be useful for various tasks including ECG classification \citep{malhotra2017timenet}. In this work, we consider application of TimeNet to phenotyping and in-hospital mortality tasks for multivariate clinical time series classification.
Deep Patient \citep{miotto2016deep} proposes leveraging features from a pre-trained stacked-autoencoder for EHR data.
However, it does not leverage the temporal aspect of the data and uses a non-temporal model based on stacked-autoencoders. 
Our approach extracts temporal features via TimeNet incorporating the sequential nature of EHR data.
Doctor AI \citep{choi2016doctor} uses discretized medical codes (e.g. diagnosis, medication, procedure) from longitudinal patient visits via a purely supervised setting while we use real-valued time series.
While approaches like Doctor AI require training a deep RNN from scratch, our approach leverages a general-purpose RNN for feature extraction.

\citep{harutyunyan2017multitask} consider training a deep RNN model for multiple prediction tasks simultaneously including phenotyping and in-hospital mortality to learn a general-purpose deep RNN for clinical time series. 
They show that it is possible to train a single network for multiple tasks simultaneously by capturing generic features that work across different tasks.
We also consider leveraging generic features for clinical time series but using an RNN that is pre-trained on diverse time series across domains, making our approach more efficient. Further, we provide an approach to rank the raw input features in order of their relevance that helps validate the models learned.

\textit{Transfer Learning via Fine-tuning}
Unsupervised pre-training has been shown to be effective in capturing the generic patterns and distribution from EHR data \citep{miotto2016deep}. 
Further, RNNs for time series classification from EHR data have been successfully explored, e.g. in \citep{lipton2015learning,che2016recurrent}.
However, these approaches do not address the challenge posed by limited labeled data, which is the focus of this work.
Transfer learning using deep neural networks has been recently explored for medical applications: 
A model learned from one hospital could be adapted to another hospital for same task via recurrent neural networks \citep{choi2016doctor}.
A deep neural network was used to transfer knowledge from one dataset to another while the source and target tasks (named-entity recognition from medical records) are the same in \citep{lee2017transfer}.
However, in both these transfer learning approaches, the source and target tasks are the same while only the dataset changes. 

In this work, we provide an approach to transfer the model trained on several healthcare-specific tasks to a different (although related) classification task using RNNs for clinical time series.
Training a deep RNN for multiple related tasks simultaneously on clinical time series has been shown to improve the performance for all tasks \citep{harutyunyan2017multitask}.
In this work, we additionally demonstrate that a model trained in this manner serves as a good initializer for building models for new related tasks.

\section{Background: TimeNet}\label{sec:bg}
Deep (multi-layered) RNNs have been shown to perform hierarchical processing of time series with different layers tackling different time scales \citep{hermans2013training,p:lstm-ad}.
TimeNet \citep{malhotra2017timenet} is a general-purpose multi-layered RNN trained on large number of diverse univariate time series from UCR Time Series Archive \citep{UCRArchive} that has been shown to be useful as off-the-shelf feature extractor for time series.
TimeNet has been trained on 18 different datasets simultaneously via an RNN autoencoder in an unsupervised manner for reconstruction task.
Features extracted from TimeNet have been found to be useful for classification task on 30 datasets not seen during training of TimeNet, proving its ability to provide meaningful features for unseen datasets.

\begin{figure*}[th]
	\centering
	\subfigure[\label{fig:ED1}]{\includegraphics[trim={10cm 1cm 25cm 0cm},clip,scale=0.25]{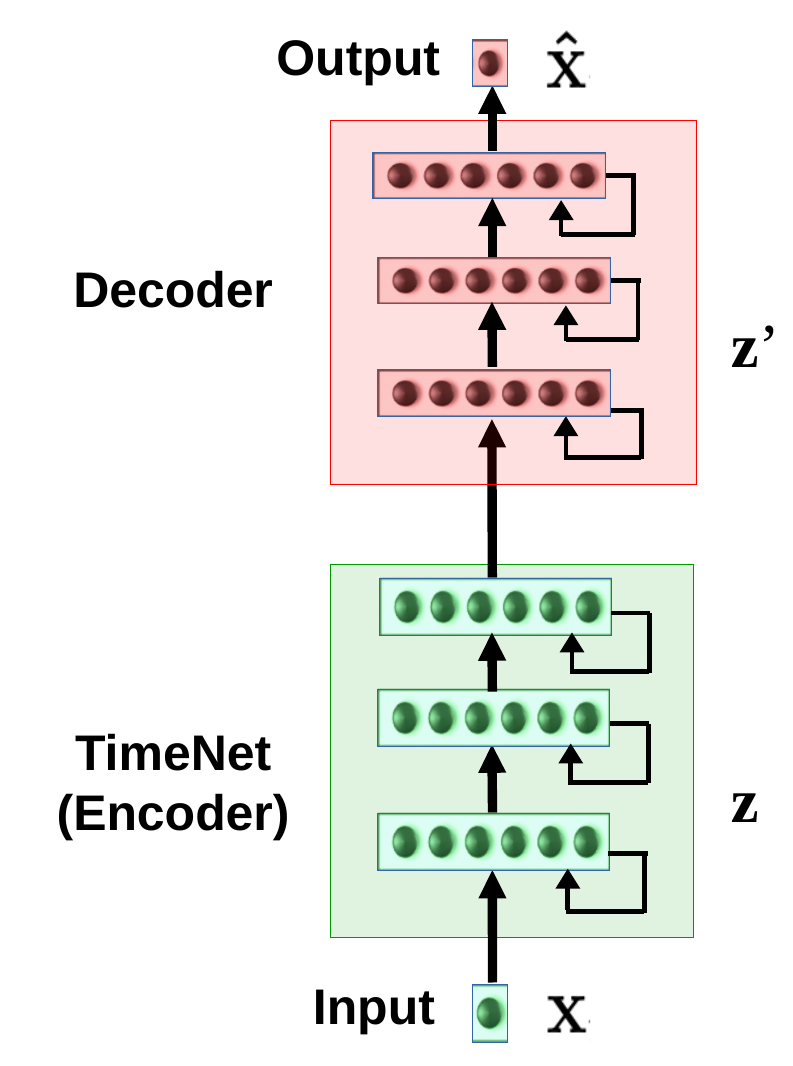}}~~
	\subfigure[\label{fig:ED-2}]{\includegraphics[width=0.65\textwidth,trim={1cm 1cm 3.8cm 3cm},scale=0.6,clip]{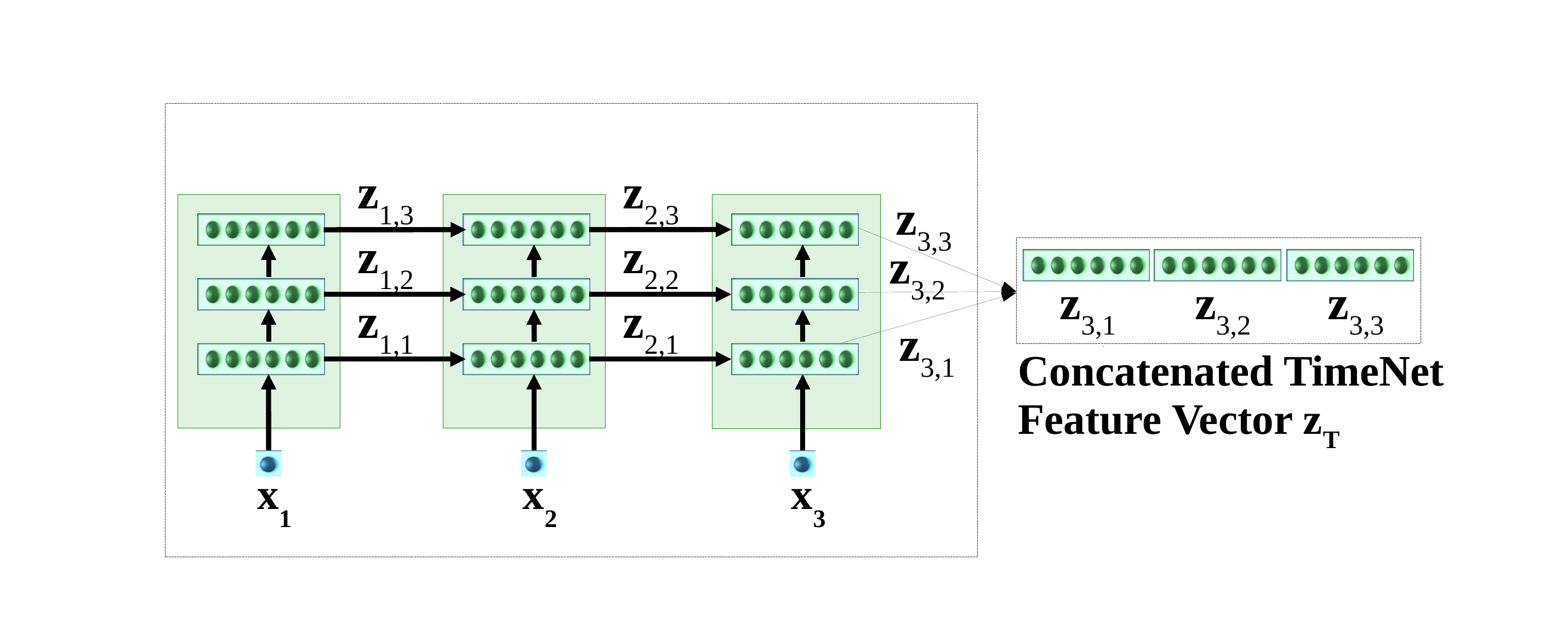}}~~
	\caption{\label{fig:ED}(a) TimeNet trained via RNN Encoder-Decoder with three hidden GRU layers, (b) TimeNet based Feature Extraction. TimeNet is shown unrolled for L = 3.}
\end{figure*}

TimeNet contains three recurrent layers having 60 Gated Recurrent Units (GRUs) \citep{cho2014learning} each.
TimeNet is an RNN trained via an autoencoder consisting of an encoder RNN and a decoder RNN trained simultaneously using the sequence-to-sequence learning framework \citep{sutskever2014sequence,bahdanau2014neural} as shown in Figure \ref{fig:ED}.
RNN autoencoder is trained to obtain the parameters $\mathbf{W}_E$ of the encoder RNN $f_E$ 
via reconstruction task such that for input $x_{1\ldots \tau}=x_1,x_2,...,x_\tau$ ($x_i \in \mathbb{R}$), the target output time series $x_{\tau\ldots 1}=x_\tau,x_{\tau-1},...,x_1$ is reverse of the input.

The RNN encoder $f_E$ provides a non-linear mapping of the univariate input time series to a fixed-dimensional vector representation $\mathbf{z}_{\tau}$ : $\mathbf{z}_{\tau}=f_E(x_{1\ldots \tau};\mathbf{W}_E)$, followed by an RNN decoder $f_D$ based non-linear mapping of $\mathbf{z}_{\tau}$ to univariate time series: $\hat{x}_{\tau\ldots 1}=f_{D}(\mathbf{z}_{\tau};\mathbf{W}_D)$; 
where $\mathbf{W}_E$ and $\mathbf{W}_D$ are the parameters of the encoder and decoder, respectively.
The model is trained to minimize the average squared reconstruction error.
%given by $\frac{1}{N\times T}\sum_{i=1}^{N}\sum_{t=1}^{T}(x_{t}^{(i)}-{\hat{x}}_{t}^{(i)})^2$, where N is the number of training instances.
%Refer \citep{malhotra2017timenet} for further details.
Training on 18 diverse datasets simultaneously results in robust time series features getting captured in $\mathbf{z}_\tau$: the decoder relies on $\mathbf{z}_\tau$ as the only input to reconstruct the time series, forcing the encoder to capture all the relevant information in the time series into the fixed-dimensional vector $\mathbf{z}_\tau$. 
This vector $\mathbf{z}_\tau$ is used as the feature vector for input $x_{1\ldots \tau}$.
This feature vector is then used to train a simpler classifier (e.g. SVM, as used in \citep{malhotra2017timenet}) for the end task.
TimeNet maps a univariate input time series to 180-dimensional feature vector, where each dimension corresponds to final output of one of the 60 GRUs in the 3 recurrent layers.

\section{Approach Overview} \label{ssec:approach_overview}
%We consider leveraging deep neural network (we take deep RNN as an example) for various classification tasks such as patient phenotyping and mortality prediction given multivariate time series corresponding to physiological parameters such as blood glucose levels, oxygen saturation levels, etc.

%In this work, we consider a \textit{domain} to be represented by a set of parameters, while a \textit{task} is represented by the final training target for the machine learning model. For example, data from same domain means that the parameters being used are same
        
Consider sets $\mathcal{D}_S$ and $\mathcal{D}_T$ of time series instances corresponding to source ($S$) and target ($T$) dataset, respectively.
$\mathcal{D}_S=\{(\mathbf{x}_S^{(i)},\mathbf{y}_S^{(i)})\}_{i=1}^{N_S}$, where $N_S$ is the number of time series instances in the source dataset. 
Denoting time series $\mathbf{x}^{(i)}_S$ by $\mathbf{x}$ and the corresponding target label $\mathbf{y}_S^{(i)}$ by $\mathbf{y}$ for simplicity of notation, we have $\mathbf{x}=\mathbf{x}_1\mathbf{x}_2\ldots\mathbf{x}_{\tau}$ denote a time series of length $\tau$, where $\mathbf{x}_t\in \mathbb{R}^n$ is an $n$-dimensional vector corresponding to $n$ parameters.
Further, $\mathbf{y}=[y_1,\ldots,y_K] \in \{0,1\}^K$, where $K$ is the number of binary classification tasks. 
Similarly, $\mathcal{D}_T=\{(\mathbf{x}_T^{(i)},y_T^{(i)})\}_{i=1}^{N_T}$ such that $N_T \ll N_S$ ,
%(i.e. the number of training instances for target task is significantly smaller than the number of training instances ),
 and $y_T^{(i)}\in\{0,1\}$ such that the target task is a binary classification task.
We consider $\mathcal{D}_S$ and $\mathcal{D}_T$ to be from same (different) \textit{domain} if the $n$ parameters in $\mathcal{D}_S$ and $\mathcal{D}_T$ are the same (different). 
%corresponding to $n$ parameters corresponds to same $n$ parameters as $\mathbf{x}_S$ in $\mathcal{D}_S$. % e.g. heart rate, pulse rate etc. are the parameters in healthcare (EHR) domain.
Further, we consider the $tasks$ for $\mathcal{D}_S$ and $\mathcal{D}_T$ to be the same if number of target classes in $\mathbf{y}_T$ and $\mathbf{y}_S$ are equal and corresponding classes are semantically same e.g. both $\mathbf{y}_S$ and $\mathbf{y}_T$ corresponding to two classes \{patient survives, patient dies\}. %and $\mathbf{y}_T$ corresponding to the same classes.

We consider two scenarios for transfer learning using RNNs\footnote{This work consolidates and extends our previous works in \citep{gupta2018using} and \citep{gupta2018using2}.}: i) \textit{domain adaptation}, $\mathcal{D}_S$ contains time series from various domains such as electric devices, motion capture, spectrographs, sensor readings, ECGs, simulated time series, etc taken from publicly available UCR Time Series Classification Archive \citep{UCRArchive}, and $\mathcal{D}_T$ contains clinical time series from EHR database (set $\mathcal{D}_T$ and $\mathcal{D}_S$ are from different domain).
We consider pre-training RNN model using $\mathcal{D}_S$ via unsupervised learning, which can provide useful features for time series from unseen domain (healthcare in our case). We adapt pre-trained model using $\mathcal{D}_T$ via supervised learning (Note: As we adapt model trained via unsupervised learning for supervised task, set $\mathcal{D}_T$ and $\mathcal{D}_S$ are from different task).
%Training a deep network on time series from various domain can provide useful features for time series from unseen domain (healthcare in our case).
ii) \textit{task adaptation}, $\mathcal{D}_S$ and $\mathcal{D}_T$ contain time series from same domain i.e. healthcare, $\mathbf{x}_S$ and $\mathbf{x}_T$ corresponding to same $n$ physiological parameters e.g. heart rate, pulse rate, oxygen saturation, etc.
Further, $\mathbf{y}_S$ corresponds to various tasks, such as presence/absence of phenotypes e.g. acute cerebrovascular disease, diabetes mellitus with complications, gastrointestinal hemorrhage, etc., and $\mathbf{y}_T$ corresponds to a related but different task e.g. present/absence of new phenotypes that are not present in $\mathbf{y}_S$ and mortality prediction. 
We consider pre-training an RNN model using $\mathcal{D}_S$ via supervised learning on a diverse set of tasks such that the model learns to capture and extract a rich set of generic features from time series that can be useful for other tasks in same domain. 
We adapt pre-trained model using $\mathcal{D}_T$ via supervised learning.\\
%Pre-training a deep networks on various tasks from same domain can learn a rich set of generic features that can be applied to a wide range of other similar tasks in same domain.
%We assume $\mathbf{y}_T$ belongs to different target labels than $\mathbf{y}_S$ e.g. mortality prediction.\\
%We consider two approaches to train models for clinical time series via transfer learning and use of pre-trained deep neural networks.}, that we describe in next two sections.

\section{Domain-adaptation: adapting universal time series feature extractors to healthcare domain}\label{ssec:approach1}

\begin{figure}[t]
	\centering
	%\captionsetup{justification=centering}
	\includegraphics[trim={1cm 10cm 1cm 2cm},clip,scale=0.5]{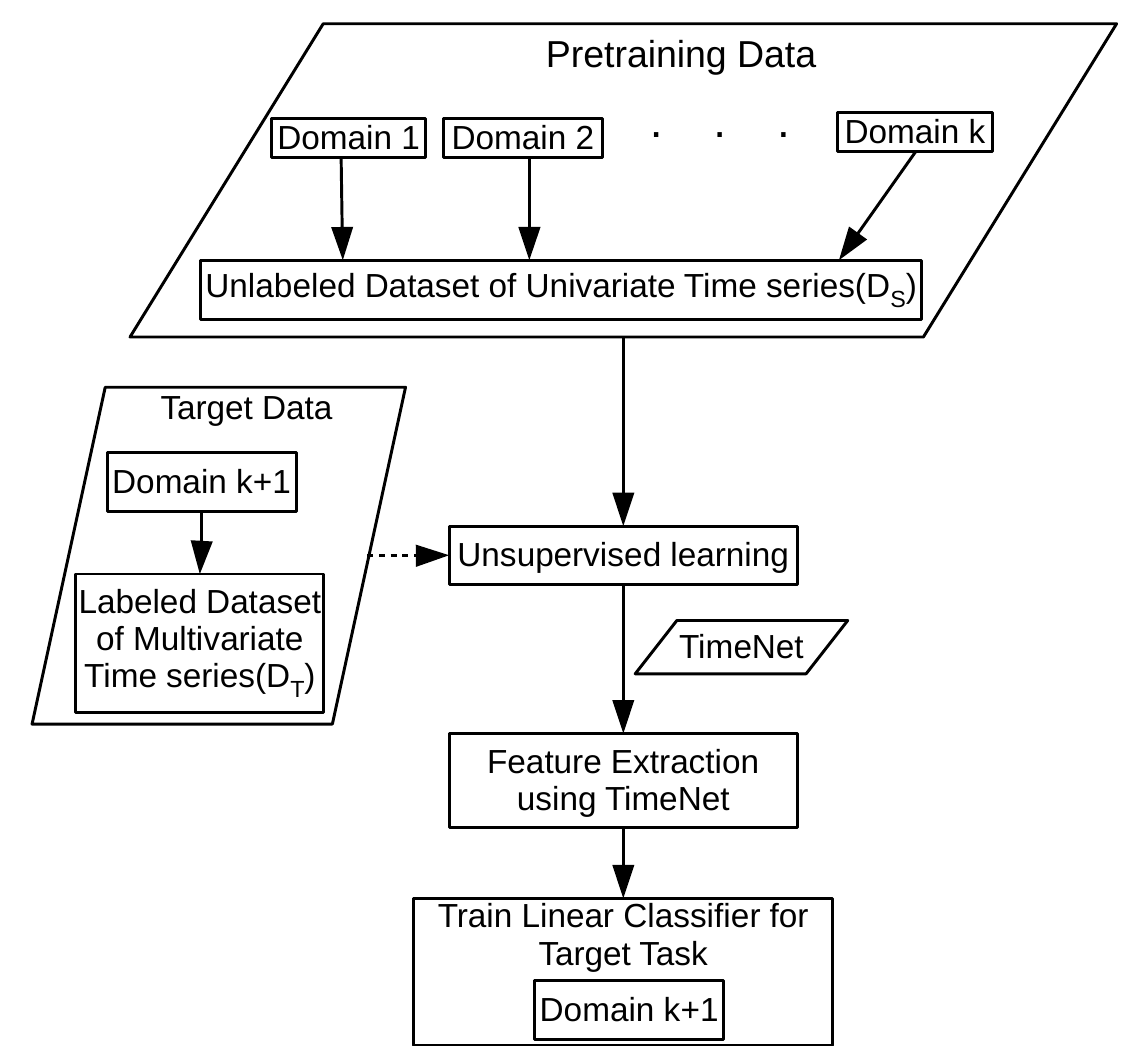}
	\caption{Domain-adaptation Scenario. A deep RNN (TimeNet) is pre-trained in an unsupervised manner on time series from $k$ diverse domains, and then used for time series feature extraction from $(k+1)$-th domain and subsequent classification. \label{fig:domain_adp}}
\end{figure}

General-purpose time series feature extractors such as TimeNet and Universal Encoder 
\citep{serra2018} usually constrain the input time series to be univariate as it is difficult to cater to multivariate time series with varying dimensionality in a single neural network. 
In this scenario, we consider adapting TimeNet to healthcare domain with two key considerations\footnote{It is to be noted that we take TimeNet as an example to illustrate our proposed domain-adaptation approach, but the proposed approach is generic 
	and can be used to adapt any universal time series feature extractor to healthcare domain.}: 
1) use TimeNet that caters to univariate time series for multivariate clinical time series which requires simultaneous consideration of various physiological parameters,
2) adapt the features from TimeNet for specific tasks from healthcare such as patient phenotyping and mortality prediction 
tasks. We show how TimeNet can be adapted to these classification tasks by training computationally efficient traditional linear classifiers on top of features extracted from TimeNet as shown in Figure \ref{fig:domain_adp}. Further, we propose a 
simple mechanism to leverage the weights of the trained linear classifier to \textit{provide insights into the relevance of each raw input feature (physiological parameter) for a given phenotype} (described in Section \ref{ssec:interpret}). 

Consider $\mathcal{D}_T$ is set of labeled time series instances from an EHR database: $\mathcal{D}_T=\{(\mathbf{x}_T^{(i)},y_T^{(i)})\}_{i=1}^{N_T}$, where $\mathbf{x}_T^{(i)}$ is a multivariate time series, 
%$y_T^{(i)} \in \{y_1,\ldots, y_C\}$, $C$ is the number of classes, $N_T$ is the number of time series instances corresponding to patients.
$y_T^{(i)} \in \{0, 1\}$ such that the target task is a binary classification task, $N_T$ is the number of time series instances corresponding to patients.
%We consider presence or absence of a phenotype as a binary classification task such that $C=2$.
We consider presence or absence of a phenotype as a binary classification task, and 
learn an independent model for each phenotype (unlike \cite{harutyunyan2017multitask} which consider phenotyping as a multi-label classification problem).
This allows us to build simple and compute-efficient linear binary classification models as described next in Section \ref{sec:pc}. 
In practice, the outputs of these binary classifiers can then be considered together to estimate the set of phenotypes present in a patient.
Similarly, mortality prediction is considered to be a binary classification task where the goal is to classify whether the patient will survive (after admission to ICU) or not.

\subsection{Feature Extraction for Multivariate Clinical Time Series}\label{ssec:learn_enc}
\begin{figure}[t]
	\centering
	\includegraphics[width=0.8\textwidth,trim={0.2cm 1.5cm 0.5cm 1cm},scale=0.1,clip]{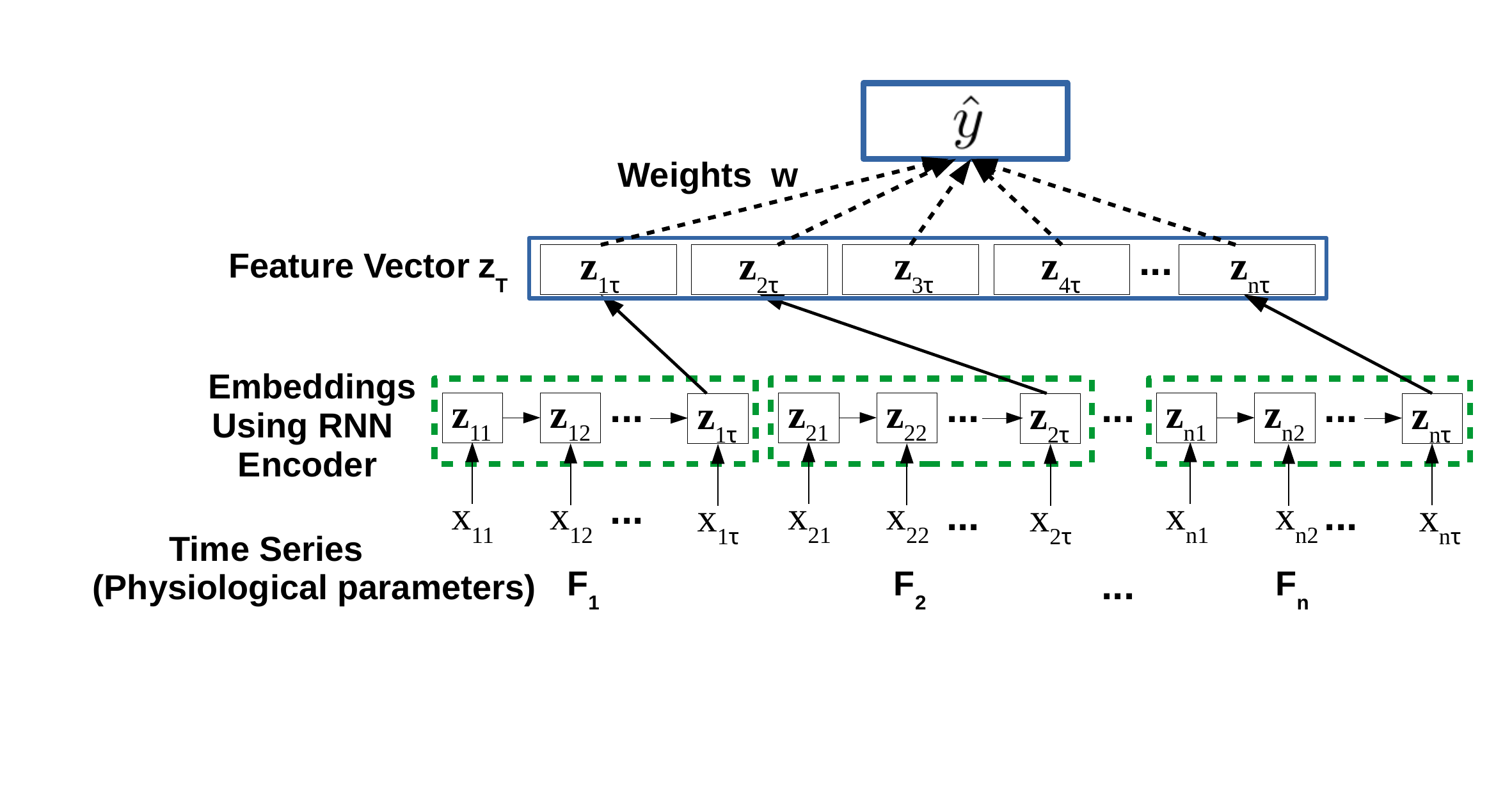}
	\caption{TimeNet based Feature Extraction and Classification.\label{fig:approach-1}}
\end{figure}

For a multivariate time series $\mathbf{x}=\mathbf{x}_1\mathbf{x}_2\ldots\mathbf{x}_{\tau}$, where $\mathbf{x}_t\in \mathbb{R}^n$, we consider time series for each of the $n$ raw input features (physiological parameters, e.g. glucose level, heart rate, etc.) independently, to obtain univariate time series $x_j={x}_{j1}{x}_{j2}\ldots{x}_{j\tau}$, $j~=1\ldots n$.
(Note: We use $\mathbf{x}$ instead of $\mathbf{x}^{(i)}$ and omit superscript $(i)$ for ease of notation).
We obtain the vector representation $\mathbf{z}_{j\tau} = f_E(x_j;\mathbf{W}_E$) for $x_j$, where $\mathbf{z}_{j\tau} \in \mathbb{R}^c$ using TimeNet as $f_E$ with $c=180$ (as described in Section \ref{sec:bg}).
In general, time series length $\tau$ also depends on $i$, e.g. based on length of stay in hospital. We omit this for sake of clarity without loss of generality. 
In practice, we convert each time series to have equal length $\tau$ by suitable pre/post-padding with 0s.
We concatenate the TimeNet-features $\mathbf{z}_{j\tau}$ for each raw input feature $j$  to get the final feature vector $\mathbf{z}_{\tau}=[\mathbf{z}_{1\tau},\mathbf{z}_{2\tau},\ldots, \mathbf{z}_{n\tau}]$ for time series $\mathbf{x}$, where $\mathbf{z}_{\tau} \in \mathbb{R}^{m}$, $m=n\times c$ as illustrated in Figure \ref{fig:approach-1}.

\subsection{Using TimeNet-based Features for Classification}\label{sec:pc}

The final concatenated feature vector $\mathbf{z}_{\tau}$ is used as input for the phenotyping and mortality prediction classification tasks. 
We note that since $c=180$ is large, $\mathbf{z}_{\tau}$ has large number of features $m\geq 180$.
We consider a linear mapping from input TimeNet features $\mathbf{z}_\tau$ to the target label $y$ s.t. the estimate $\hat{y}=\mathbf{w}\cdot \mathbf{z}_\tau$, where $\mathbf{w}\in \mathbb{R}^m$. 
We constrain the linear model with weights $\mathbf{w}$ to use only a few of these large number of features. 
The weights are obtained using LASSO-regularized loss function \cite{tibshirani1996regression}:
\begin{equation}\label{eq:lasso}\argminA_{\mathbf{w}}
\frac{1}{N}\sum_{i=1}^{N}(y^{(i)}-\mathbf{w}\cdot\mathbf{z}_\tau^{(i)})^2 + \alpha||\mathbf{w}||_1
\end{equation}
where $y^{(i)}\in \{0,1\}$, $||\mathbf{w}||_1=\sum_{j=1}^n\sum_{k=1}^c|w_{jk}|$ is the L$_1$-norm, where $w_{jk}$ represents the weight assigned to the $k$-th TimeNet-feature for the $j$-th raw feature, and $\alpha$ controls the extent of sparsity -- with higher $\alpha$ implying more sparsity, i.e. fewer TimeNet features are selected for the final classifier.
%This approach selects relevant TimeNet features for the end task, and uses a linear combination of selected features to estimate the class.

\begin{figure}[ht]
	\centering
	\includegraphics[width=0.6\textwidth,trim={0.cm 2cm 13.5cm 2cm},clip]{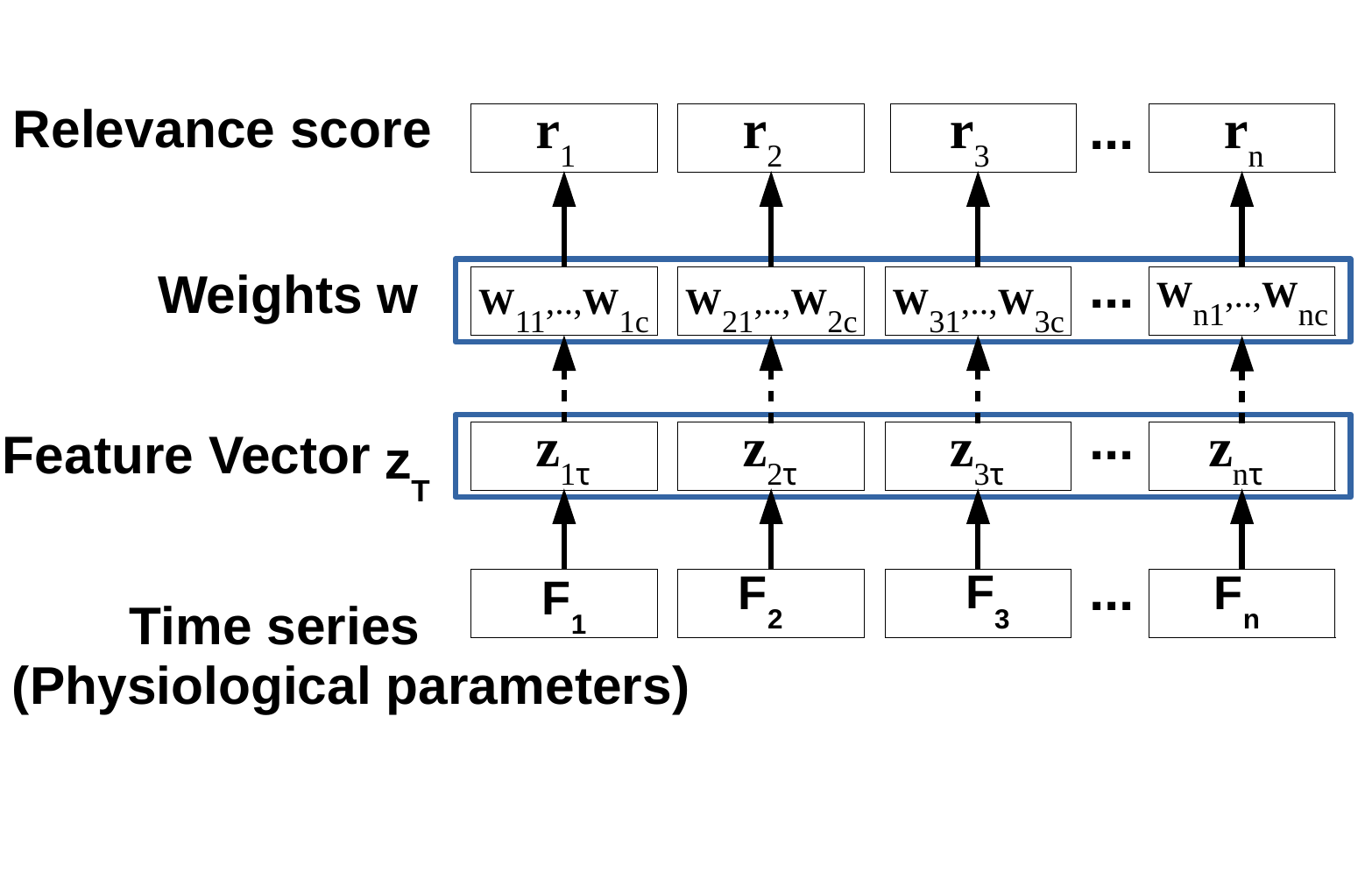}
	\caption{Obtaining relevance scores for raw input features. Here, $\tau$: time series length, $n$: number of raw input features.\label{fig:rel_1}}
\end{figure}

\subsection{Obtaining Relevance Scores for Raw Features\label{ssec:interpret}}
Determining relevance of the $n$ raw input features for a given phenotype is potentially useful to obtain insights into the obtained classification model. 
The sparse weights $\mathbf{w}$ are easy to interpret and can give interesting insights into relevant features for a classification task (e.g. as used in \cite{micenkova2013explaining}). 
We obtain the relevance $r_j$ of the $j$-th raw input feature as the sum of the absolute values of the weights $w_{jk}$ assigned to the corresponding TimeNet features $\mathbf{z}_{j\tau}$ as shown in Figure \ref{fig:rel_1}, s.t.
\begin{equation}\label{eq:relevance}
r_j = \sum_{k=1}^c|w_{jk}|, ~ j=1\ldots n.
\end{equation}
Further, $r_j$ is normalized using min-max normalization such that $r'_j = \frac{r_j - r_{min}}{r_{max}-r_{min}}\in [0,1]$; $r_{min}$ is minimum of $\{r_1,\ldots,r_n\}$, $r_{max}$ is maximum of $\{r_1,\ldots,r_n\}$.
In practice, this kind of relevance scores for the raw features help to interpret and validate the overall model. For example, one would expect blood glucose level feature to have a high relevance score when learning a model to detect diabetes mellitus phenotype (we provide such insights later in Section \ref{ssec:timenet}).
%Higher the relevance score for a raw feature more is the significance of raw feature in final decision of classifier.  

\section{Task-adaptation: adapting healthcare-specific pre-trained models to a new task}\label{ssec:approach2}

\begin{figure}[t]
	\centering
	%\captionsetup{justification=centering}

	\includegraphics[trim={1cm 9.5cm 1cm 1cm},clip,scale=0.5]{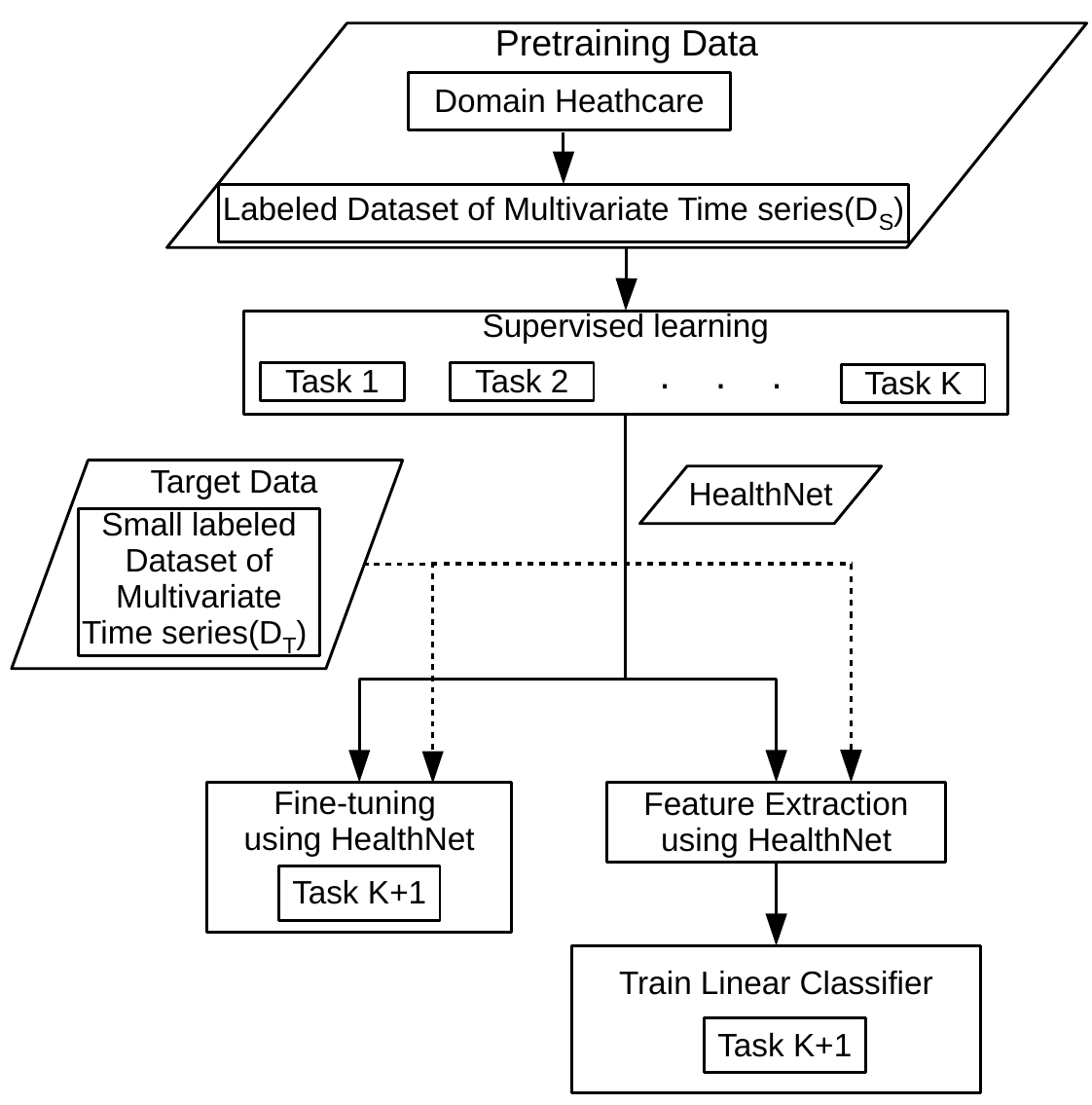}
	\caption{Task-adaptation Scenario. A deep RNN (HealthNet) is pre-trained for $K$ classification tasks from healthcare domain simultaneously via supervised training. Then, it is adapted for target task either via fine-tuning or feature extraction (potentially using only a small amount of labeled training data for target task). \label{fig:task_adp}}
\end{figure}

\begin{figure*}[ht]
	\centering
	\subfigure[\label{fig:HealthNet}]{\includegraphics[trim={10cm 0.5cm 25cm 8cm},clip,scale=0.35]{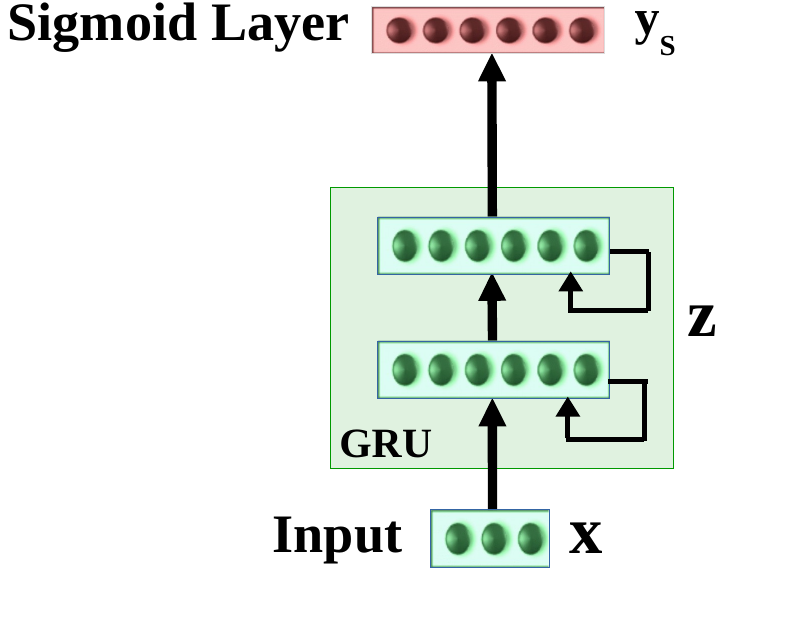}}~~
	\subfigure[\label{fig:finetune}]{\includegraphics[width=0.65\textwidth,trim={9.2cm 3.6cm 17.5cm 2cm},scale=0.1,clip]{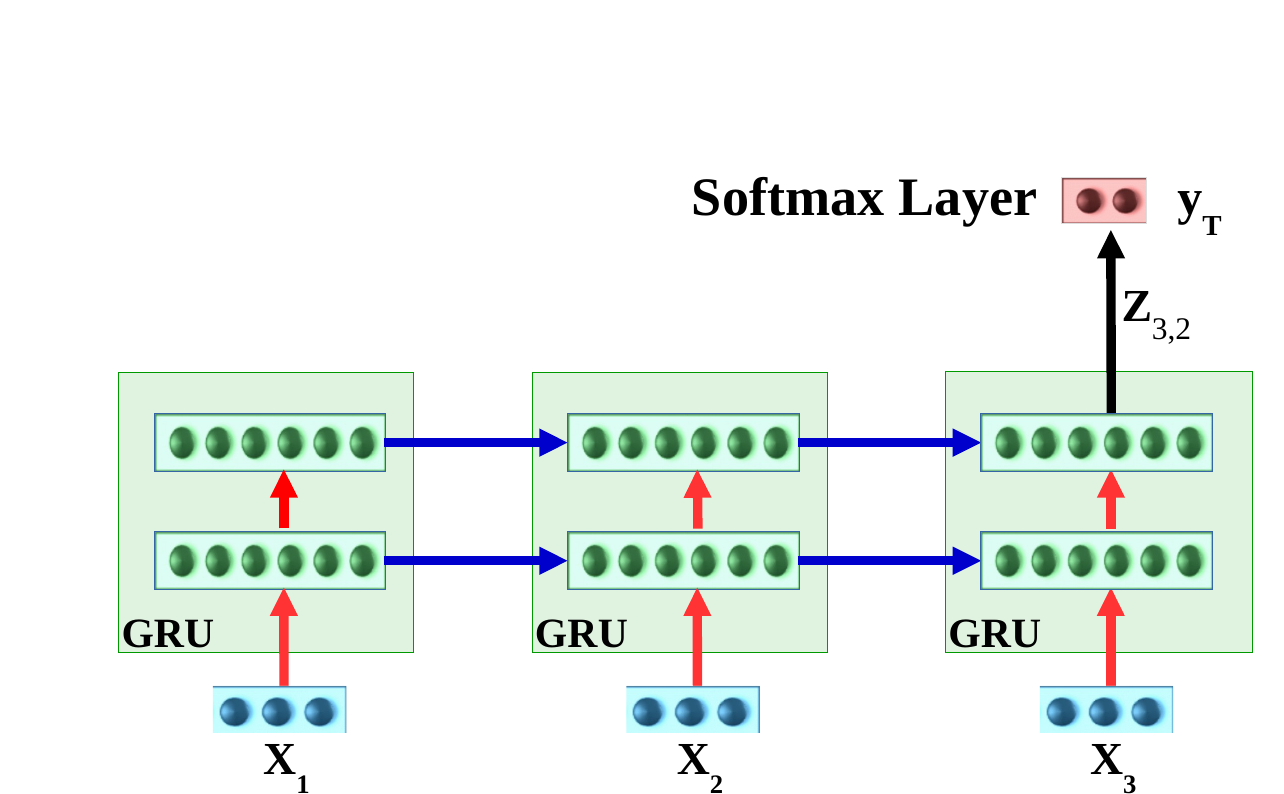}}~~
	\caption{\label{fig:approach-2} (a) HealthNet trained via supervised learning for multiple source tasks simultaneously using final activation as sigmoid, (b) fine-tuning HealthNet for a new target task using 
	final activation as softmax. Here, blue and red arrows corresponds to recurrent and feed forward weights of the recurrent layers respectively. Only feed forward (red) weights are regularized while fine-tuning the HealthNet. RNN with L = 2 hidden layers is shown unrolled over $\tau$ = 3 time steps.} 
\end{figure*}

\begin{figure}[th]
	\centering
	\includegraphics[trim={1cm 0.4cm 10.2cm 5cm},clip,width=\columnwidth]{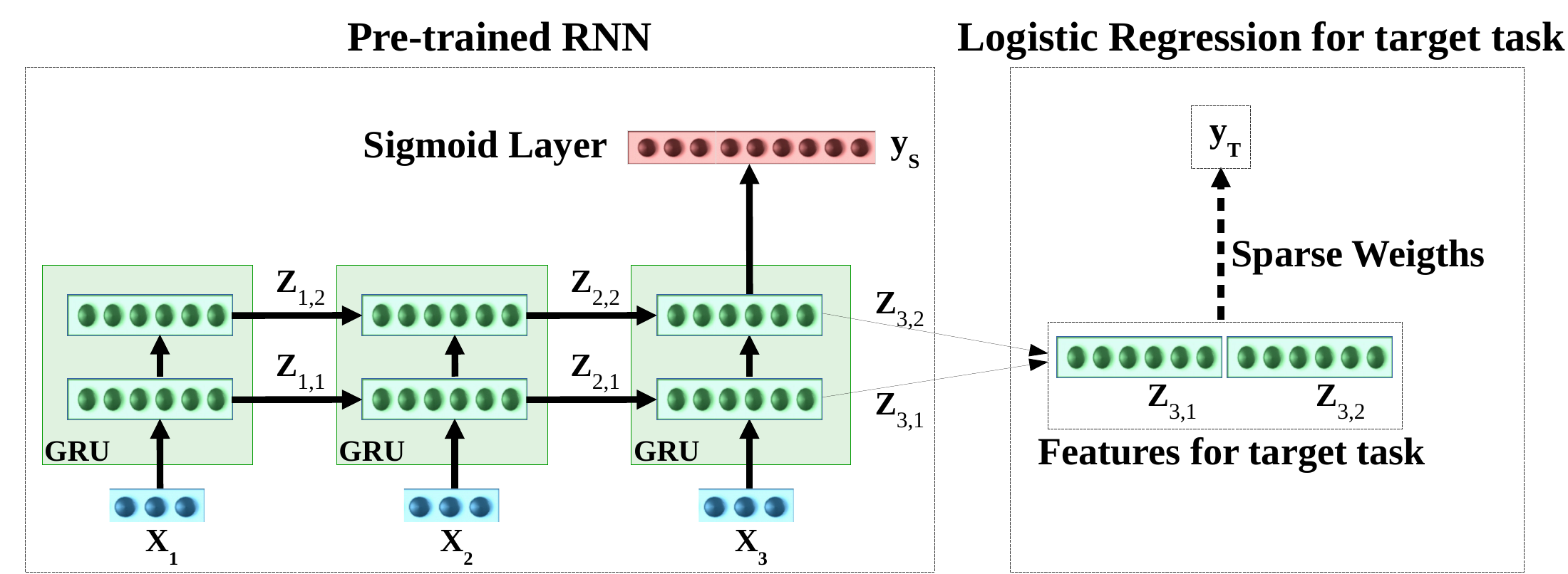}
	%\vspace{-10pt}
	\caption{\label{fig:multilabel} Inference in Task-Adaptation: Using features extracted from HealthNet. RNN with $L=2$ hidden layers is shown unrolled over $\tau=3$ time steps.}
\end{figure}

In this scenario, the goal is to transfer the learning from a set of tasks to another related task for clinical time series by means of an RNN. 
Considering phenotype detection from time series of physiological parameters as a binary classification task, we train HealthNet as an RNN classifier on a diverse set of such binary classification tasks (one task per phenotype) simultaneously using 
a large labeled dataset. We consider following approaches to adapt it to an unseen target task as shown in Figure \ref{fig:task_adp}: 
\begin{enumerate}
	\item We initialize parameters of the target task specific RNN using the parameters of HealthNet previously trained on a large number of source tasks;  
	so that HealthNet provides good initialization of parameters of task-specific RNN and train the model (described in Section \ref{ssec:HN-L1/L2}). 
	\item We extract features using HealthNet and then train an easily trainable non-temporal linear classification model such as a logistic regression model \citep{hosmer2013applied} for target tasks, i.e. 
	identifying a new phenotype and predicting in-hospital mortality, with few labeled instances (described in Section \ref{ssec:LR}).
\end{enumerate}

More specifically, consider $\mathcal{D}S$ and $\mathcal{D}_T$ are sets of labeled time series instances from an EHR database (i.e. same $domain$):
$\mathcal{D}_S=\{(\mathbf{x}_S^{(i)},\mathbf{y}_S^{(i)})\}_{i=1}^{N_S}$, where $\mathbf{x}_S^{(i)}$ is a multivariate time series, % corresponding to $n$ physiological parameters such as glucose level, heart rate, etc, 
$\mathbf{y}_S^{(i)} = \{y_1,\ldots, y_K\} \in \{0,1\}^K$, $K$ is the number of binary classification tasks, $N_S$ is the number of time series instances corresponding to patients.
%$\mathcal{D}_S=\{(\mathbf{x}_S^{(i)},\mathbf{y}_S^{(i)})\}_{i=1}^{N_S}$, where $N_S$ is the number of time series instances corresponding to patients. %$N_S$ patients %(in our experiments, we consider each episode of hospital stay for a patient as a separate data instance). 
%Denoting time series $\mathbf{x}^{(i)}_S$ by $\mathbf{x}$ and the corresponding target label $\mathbf{y}_S^{(i)}$ by $\mathbf{y}$ for simplicity of notation, we have $\mathbf{x}=\mathbf{x}_1\mathbf{x}_2\ldots\mathbf{x}_{\tau}$ denote a time series of length $\tau$, where $\mathbf{x}_t\in \mathbb{R}^n$ is an $n$-dimensional vector corresponding to $n$ physiological parameters such as glucose level, heart rate, etc.
%Further, $\mathbf{y}=[y_1,\ldots,y_K] \in \{0,1\}^K$, where $K$ is the number of binary classification tasks. 
%For example, for $K=5$ binary classification tasks corresponding to presence or absence of 5 phenotypes, $\mathbf{y}=[1,0,1,1,0]$ indicates that phenotypes 1, 3, and 4 are present while phenotypes 2 and 5 are absent.
Similarly, $\mathcal{D}_T=\{(\mathbf{x}_T^{(i)},y_T^{(i)})\}_{i=1}^{N_T}$ such that $N_T \ll N_S$, and $y_T^{(i)}\in\{0,1\}$ such that the target task is a binary classification task.
%We assume that the time series in $\mathcal{D}_T$ belongs to same $n$ physiological parameters as in $\mathcal{D}_S$.
We first train HealthNet on $K$ source tasks using $\mathcal{D}_S$ (refer Section \ref{ssec:train_HealthNet} for details), as shown in Figure \ref{fig:HealthNet}, and then consider following two scenarios for adapting to the target tasks using $\mathcal{D}_T$: 
\begin{itemize}
        \item fine-tuning the HealthNet with suitable regularization (refer Section \ref{ssec:HN-L1/L2} for details), as shown in Figure \ref{fig:finetune}. This allow us to train model that does not require 
        hyper-parameter tuning efforts.
	\item train the simpler logistic regression (LR) classifier for target task using $\mathcal{D}_T$ and the features obtained via HealthNet (refer Section \ref{ssec:LR} for details), as shown in Figure \ref{fig:multilabel}, which is compute-efficient.

\end{itemize}
We next provide details of training RNN and LR models.

\iffalse
\begin{figure}[th]
	\centering
	\includegraphics[trim={1cm 0.4cm 10.2cm 5cm},clip,width=\columnwidth]{figures/finetune.pdf}
	%\vspace{-10pt}
	\caption{\label{fig:finetune} Inference in the proposed transfer learning approach. RNN with $L=2$ hidden layers is shown unrolled over $\tau=3$ time steps.}
\end{figure}
\fi

\subsection{Obtaining HealthNet using Supervised Pre-training of RNN}\label{ssec:train_HealthNet}
Training an RNN on $K$ binary classification tasks simultaneously can be considered as a multi-label classification problem.
We train a multi-layered RNN with $L$ recurrent layers having Gated Recurrent Units (GRUs) \citep{cho2014learning} to map $\mathbf{x}^{(i)} \in \mathcal{D}_S$ to $\mathbf{y}^{(i)}$.
Let $\mathbf{z}_{t,l}\in \mathbb{R}^h$ denote the output of recurrent units in $l$-th hidden layer at time $t$, and $\mathbf{z}_{t}=[\mathbf{z}_{t,1},\ldots,\mathbf{z}_{t,L}] \in \mathbb{R}^m$ denote the hidden state at time $t$ obtained as concatenation of hidden states of all layers, where $h$ is the number of GRU units in a hidden layer and $m=h\times L$.
The parameters of the network 
are obtained by minimizing the cross-entropy loss given by $\mathcal{L}$ via stochastic gradient descent: 
\begin{equation}\label{eq:RNN-C}
\begin{aligned}
\mathbf{z}^{(i)}_\tau&=f_E(\mathbf{x}^{(i)};\mathbf{W}'_E),~
\mathbf{\hat{y}}^{(i)} = \sigma(\mathbf{W}_C\:\mathbf{z}^{(i)}_{\tau,L}+\mathbf{b}_C)\\
C(y_k^{(i)},\hat{y}_k^{(i)})&=y_k^{(i)}\cdot log(\hat{y}_k^{(i)})+(1-y_k^{(i)})\cdot log((1-\hat{y}_k^{(i)}))\\
\mathcal{L}&=-\frac{1}{N_S\times K}\sum_{i=1}^{N_S}\sum_{k=1}^{K}C(y_k^{(i)},\hat{y}_k^{(i)}).
\end{aligned}
\end{equation}
Here $\sigma({x})$ =  $({1+e^{-x}})^{-1}$ is the sigmoid activation function, $\mathbf{\hat{y}}^{(i)}$ is the estimate for target $\mathbf{y}^{(i)}$, $\mathbf{W}'_E$ are parameters of recurrent layers, and $\mathbf{W}_C$ and $\mathbf{b}_C$ are parameters of the classification layer.

\subsection{Task-Adaptation: Fine-tuning of HealthNet\label{ssec:HN-L1/L2}}

We initialized the target task specific RNN parameters by the pre-trained RNN parameters of recurrent layers ($\mathbf{W}'_E$) and a new binary classification layer parameters ($\mathbf{W}'_C$ and $\mathbf{b}'_C$).
We obtain probabilities of two classes for the binary classification task as $\mathbf{\hat{y}}^{(i)}$ = $softmax(\mathbf{W}'_C\:\mathbf{z}'^{(i)}_{\tau,L}+\mathbf{b}'_C)$, where
%$\mathbf{W}'_C$ and $\mathbf{b}'_C$ are parameters of the classification layer, 
$\mathbf{z}'^{(i)}_{\tau,L}$ is the output of recurrent units in last layer (L) at last timestamp (${\tau}$). 
Let $\mathbf{W}'_{EF}$ and $\mathbf{W}'_{ER}$ are feed forward and recurrent weights of the recurrent layers.
All parameters are trained together by minimizing cross-entropy loss with regularizer. We consider two regularizer techniques to obtain two different fine-tuned models with loss given by $\mathcal{L}_1$ and $\mathcal{L}_2$ via stochastic gradient descent:

\iffalse
Let $\mathbf{W}'_E$ are parameters of recurrent layers and $\mathbf{W}'_{EF}$ and $\mathbf{W}'_{ER}$ are feed forward and recurrent weights of the recurrent layers. 
All parameters of RNN network are trained. We obtain probability of the positive class for the binary classification task as $\hat{y}^{(i)}$ = $softmax(\mathbf{W}'_C\:\mathbf{z}'^{(i)}_{\tau,L}+\mathbf{b}'_C)$, where
$\mathbf{W}'_C$ and $\mathbf{b}'_C$ are parameters of the classification layer, $\mathbf{z}'^{(i)}_{\tau,L}$ is the output of recurrent units in last layer (L) at last timestamp (${\tau}$). Let $\mathbf{W}'_E$ are parameters of recurrent layers and $\mathbf{W}'_{EF}$ and $\mathbf{W}'_{ER}$ are feed forward and recurrent weights of the recurrent layers. 
The parameters are obtained by minimizing cross-entropy loss with regularizer.
We consider two regularizer techniques to obtain two different fine-tuned models with loss given by $\mathcal{L}_1$ and $\mathcal{L}_2$ via stochastic gradient descent:
\fi

\begin{equation}\label{eq:RNN-C-tune-L1}
\begin{aligned}
\mathcal{L}_1&=-\frac{1}{N_T}\sum_{i=1}^{N_T}C(y^{(i)},\hat{y}^{(i)})+ \lambda\mathbf{\|{W}'}_{EF}\|_1\\
\end{aligned}
\end{equation}

\begin{equation}\label{eq:RNN-C-tune-L2}
\begin{aligned}
\mathcal{L}_2&=-\frac{1}{N_T}\sum_{i=1}^{N_T}C(y^{(i)},\hat{y}^{(i)})+ \lambda\mathbf{\|{W}'}_{EF}\|_2\\
\end{aligned}
\end{equation}

where $\hat{y}^{(i)}$ is the probability of positive class,  $||\mathbf{W}'_{EF}||_1=\sum_{j=1}^m|W_{j}|$ is the $L_1$ norm with $\lambda$ controlling the extent of sparsity, and $||\mathbf{W}'_{EF}||_2=\sum_{j=1}^m {W_{j}}^2$ is the $L_2$ norm. 
As \cite{bengio2013regularizer} suggest that using an ${L}_1$ or ${L}_2$ penalty on the recurrent weights compromises the ability of the network to learn and retain information through time, 
therefore, we apply L1 or L2 regularizer only to the feed forward connections across recurrent layers and not the weights of the recurrent connections.

%suggests using an ${L}_1$ or ${L}_2$ penalty on the recurrent weights can help with exploding gradients and compromises the cells' ability to learn and retain information through time, 
%we applied regularizer to the feed forward weights across recurrent layers. 

\subsection{Task-Adaptation: Using features extracted from HealthNet \label{ssec:LR}}
For input $\mathbf{x}^{(i)} \in \mathcal{D}_T$, the hidden state $\mathbf{z}^{(i)}_{\tau}$ at last time step $\tau$ is used as input feature vector for training the LR model. 
We obtain probability of the positive class for the binary classification task as $\hat{y}^{(i)}$ = $\sigma(\mathbf{w}'_C\:\mathbf{z}^{(i)}_{\tau}+{b}'_C)$, where $\mathbf{w}'_C$, ${b}'_C$ are parameters of LR.
The parameters are obtained by minimizing the negative log-likelihood loss $\mathcal{L'}$: 

\begin{equation}\label{eq:RNN-C-tune}
\begin{aligned}
\mathcal{L'}&=-\frac{1}{N_T}\sum_{i=1}^{N_T}C(y^{(i)},\hat{y}^{(i)})+ \lambda\mathbf{\|{w}'}_C\|_1\\
\end{aligned}
\end{equation}
where $||\mathbf{w}'_C||_1=\sum_{j=1}^m|w_{j}|$ is the L$_1$ regularizer with $\lambda$ controlling the extent of sparsity -- with higher $\lambda$ implying more sparsity, i.e. fewer features from the representation vector are selected for the final classifier. 
It is to be noted that this way of training the LR model on pre-trained RNN features is equivalent to freezing the parameters of all the hidden layers of the pre-trained RNN while tuning the parameters of a new final classification layer. 
The sparsity constraint ensures that only a small number of parameters are to be tuned which is useful to avoid overfitting when labeled data is small.

\section{Dataset Description}\label{ssec:data_desc}
We use MIMIC-III (v1.4) clinical database \cite{johnson2016mimic} which consists of over 60,000 ICU stays across 40,000 critical care patients. 
We use same experimental setup as in \cite{harutyunyan2017multitask}, with same splits and features for train, validation and test datasets\footnote{https://github.com/yerevann/mimic3-benchmarks\label{footnote}} 
based on 17 physiological parameters with 12 real-valued (e.g. blood glucose level, systolic blood pressure, etc.) and 5 categorical time series (e.g. Glascow coma scale motor response, Glascow coma scale verbal, etc.), sampled at 1 hour intervals. 
The categorical variables are converted to one-hot vectors such that final multivariate time series has $n=76$ raw input features (59 actual features and 17 masking features to denote missing values).
%We use time series from only up to first $48$ hours of ICU stay for all predictions (such that $\tau=48$) to imitate the practical scenario where early predictions are important.
In all our experiments, we restrict training time series data up to first $48$ hours in ICU stay, such that $\tau=48$ while training all models to imitate practical scenario where early predictions are important, unlike \cite{harutyunyan2017multitask,song2017attend} which use entire time series for training the classifier for phenotyping task.
We consider each episode of hospital stay for a patient as a separate data instance.

The benchmark dataset contains label information for presence/absence of 25 phenotypes common in adult ICUs (e.g. acute cerebrovascular disease, diabetes mellitus with complications, gastrointestinal hemorrhage, etc.), and in-hospital mortality, whether patient survived or not after ICU admission (class 1: patient dies, class 0: patient survives).

\iffalse
The data contains multivariate time series for multiple physiological parameters with 12 real-valued (e.g. blood glucose level, systolic blood pressure, etc.) and 5 categorical parameters (e.g. Glascow coma scale motor response, Glascow coma scale verbal, etc.), sampled at 1 hour interval. 
The categorical variables are converted to one-hot vectors such that final multivariate time series has dimension $n=76$. 
We use time series from only up to first $48$ hours of ICU stay for all predictions (such that $\tau=48$) to imitate the practical scenario where early predictions are important.
\fi

\section{Experimental Evaluation of TimeNet based Transfer Learning\label{ssec:timenet}}
\begin{table*}
	\centering
	\footnotesize
	\caption{\label{tab:approach-1}Classification Performance Comparison for Phenotyping Task. Here, LR: Logistic regression, LSTM-Multi: LSTM-based multitask model, SAnD (Simply Attend and Diagnose): Fully attention-based model, SAnD-Multi: SAnD-based multitask model. (Note: For phenotyping, we compare TimeNet-48-Eps with existing benchmarks over TimeNet-All-Eps as it is more applicable in practical scenarios.)}
	\begin{tabular}{|c|c|c|c|}
		\hline
		\textbf{Approach}&\textbf{Micro AUC}&\textbf{Macro AUC}&\textbf{Weighted AUC}\\
		\hline
		\textbf{LR}&0.801&0.741&0.732\\\hline
		\textbf{LSTM}&\textbf{0.821}&0.77&0.757\\\hline
		\textbf{LSTM-Multi}&0.817&0.766&0.753\\\hline\hline		
		\textbf{SAnD}&0.816&0.766&0.754\\\hline
		\textbf{SAnD-Multi}&0.819&0.771&0.759\\\hline\hline
		\textbf{TimeNet-48}&0.812&0.761&0.751\\\hline
		\textbf{TimeNet-All}&0.813&0.764&0.754\\\hline
		\textbf{TimeNet-48-Eps}&0.820&\textbf{0.772}&\textbf{0.765}\\\hline
		\textbf{TimeNet-All-Eps}&\textit{0.822}&\textit{0.775}&\textit{0.768}\\\hline
	\end{tabular}

\end{table*}

\begin{table*}
	\centering
	\footnotesize
	\caption{\label{tab:mor_result}Performance Comparison for In-hospital Mortality Prediction Task. Here, LR: Logistic regression, LSTM-Multi: LSTM-based multitask model, SAnD (Simply Attend and Diagnose): Fully attention-based model, SAnD-Multi: SAnD-based multitask model. (Note: Only TimeNet-48 variant is applicable for in-hospital mortality task.)}
	\begin{tabular}{|c|c|c|c|}
		\hline
		\textbf{Approach}&\textbf{AUROC}&\textbf{AUPRC}&\textbf{min(Se,$^+$P)}\\
		\hline
		\textbf{LR}&0.845&0.472&0.469\\\hline
		\textbf{LSTM}&0.854&0.516&0.491\\\hline
		\textbf{LSTM-Multi}&\textbf{0.863}&0.517&0.499\\\hline\hline		
		\textbf{SAnD}&0.857&0.518&0.5\\\hline
		\textbf{SAnD-Multi}&0.859&\textbf{0.519}&\textbf{0.504}\\\hline\hline
		\textbf{TimeNet-48}&0.852&\textbf{0.519}&0.486\\\hline
	\end{tabular}
	
\end{table*}

We evaluate TimeNet based Transfer Learning approach on binary classification tasks (i) presence/absence of 25 phenotypes, and (ii) in-hospital mortality task.

\subsection{Experimental Setup}
We have $n=76$ raw input features resulting in $m=13,680$-dimensional ($m=76\times 180$) TimeNet feature vector for each admission. 
We use $\alpha=0.0001$ for phenotype classifiers and use $\alpha=0.0003$ for in-hospital mortality classifier ($\alpha$ is chosen based on hold-out validation set).
Table \ref{tab:approach-1} and Table \ref{tab:mor_result} summarize the results of phenotyping and in-hospital mortality prediction task respectively, and provides comparison with existing benchmarks.
Refer Table \ref{tab:phenotype-wise} for detailed phenotype-wise results, and Table \ref{tab:feature} for names of raw features used.

We consider two variants of classifier models for phenotyping task: i)
\textit{TimeNet-x} using data from current episode, ii) \textit{TimeNet-x-Eps} using data from previous episode of a patient as well (whenever available) via an additional input feature related to presence or absence of the phenotype in previous episode. 
Each classifier is trained using up to first 48 hours of data after ICU admission. 
However, we consider two classifier variants depending upon hours of data $x$ used to estimate the target class at test time.
For $x=48$, data up to first 48 hours after admission is used for determining the phenotype.
For $x=All$, the learned classifier is applied to all 48-hours windows (overlapping with shift of 24 hours) over the entire ICU stay period of a patient, and the average phenotype probability across windows is used as the final estimate of the target class.
In \textit{TimeNet-x-Eps}, the additional feature is related to the presence (1) or absence (0) of the phenotype during the previous episode.
We use the ground-truth value for this feature during training time, and the probability of presence of phenotype during previous episode (as given via LASSO-based classifier) at test time.
%Note: As stated earlier, for in-hospital mortality task, we only consider $x=48$ and do not test $x=All$ scenario.

\begin{figure}
	\centering
	\subfigure[P$_1$\label{fig:p1}]{\includegraphics[trim={0cm 0.5cm 1.5cm 0.5cm},clip,scale=0.15]{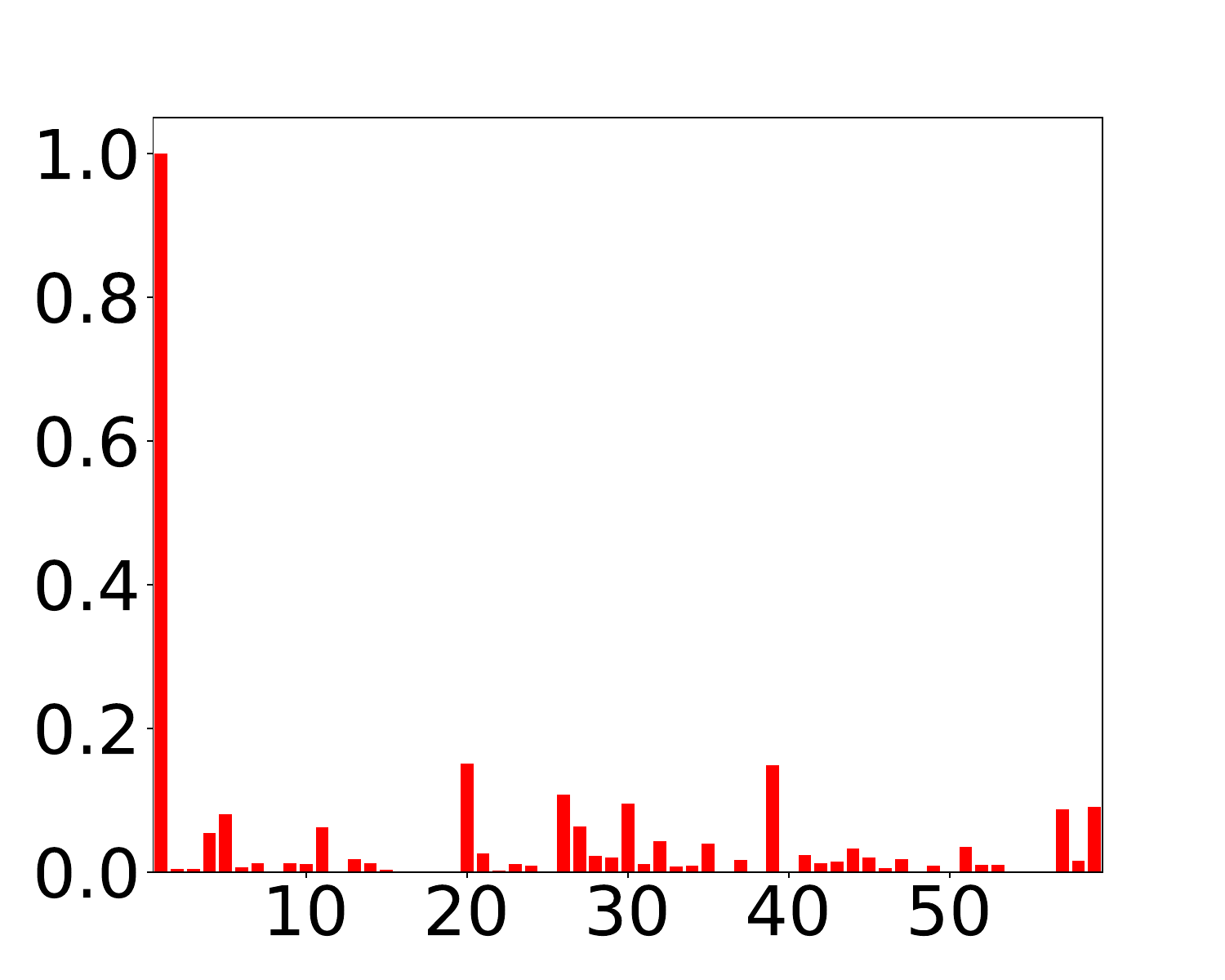}}
	\subfigure[P$_2$\label{fig:p2}]{\includegraphics[trim={0.5cm 0.5cm 1.5cm 0.5cm},clip,scale=0.15]{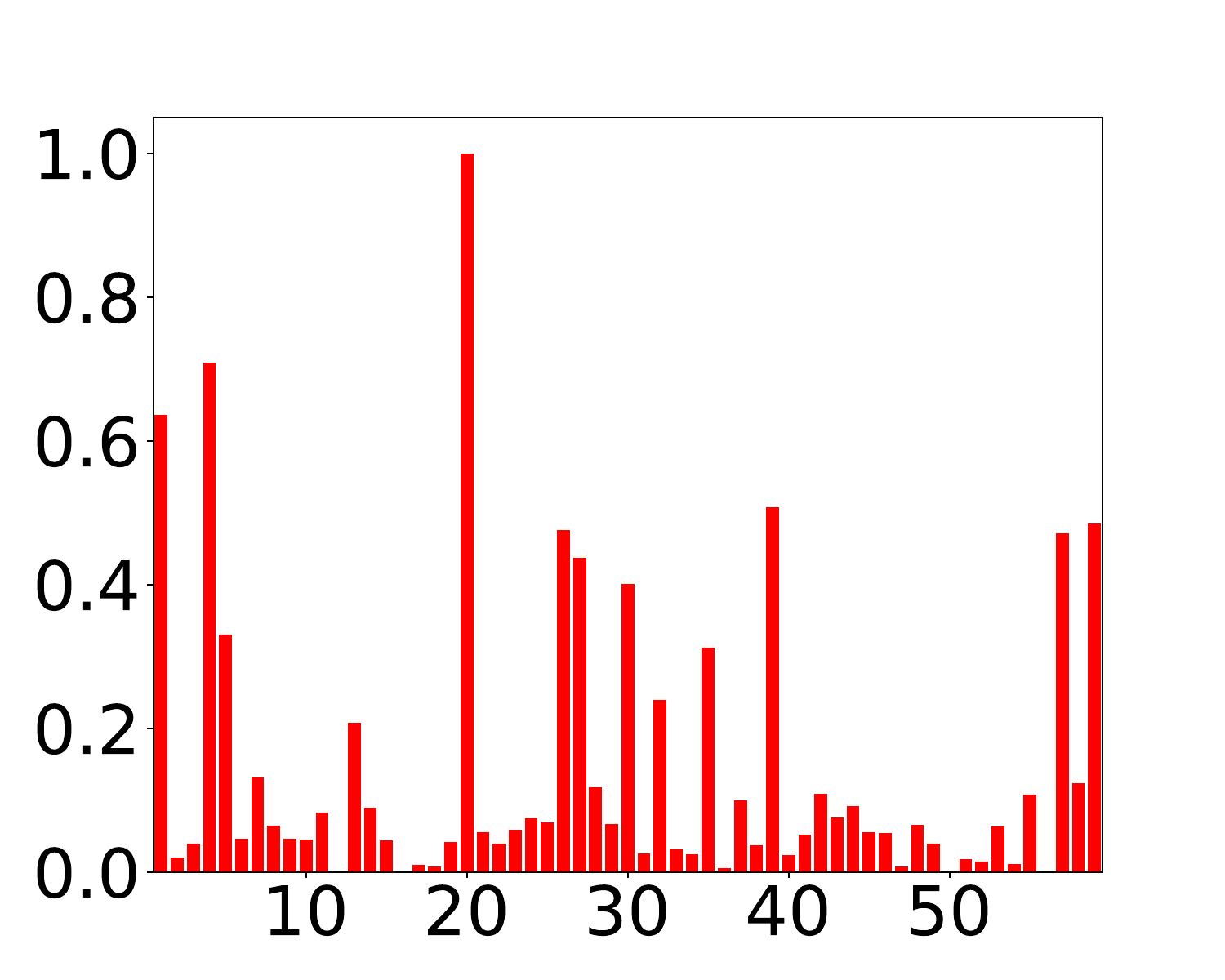}}
	\vspace{-0.3cm}
	\caption{Feature relevance after LASSO. x-axis: Feature Number, y-axis: Relevance Score. Here, P$_1$: Diabetes Mellitus with Complications, P$_2$: Essential Hypertension.\label{fig:relevance}}
\end{figure}

\subsection{Results and Observations}
\subsubsection{Classification Tasks}
For the phenotyping task, we make following observations from Table \ref{tab:approach-1}:
\newline 1. \textit{TimeNet-48 vs LR}: \textit{TimeNet-based features perform significantly better than hand-crafted features} as used in LR (logistic regression), while using first 48 hours of data only unlike the LR approach that uses entire episode's data. This proves the effectiveness of TimeNet features for MIMIC-III data. 
Further, it only requires \textit{tuning a single hyper-parameter $\alpha$} for LASSO, unlike other approaches like LSTM \cite{harutyunyan2017multitask} that would involve tuning number of hidden units, layers, learning rate, etc.
\newline 2. \textit{TimeNet-$x$ vs TimeNet-$x$-Eps}: Leveraging previous episode's time series data for a patient significantly improves the classification performance.
\newline 3. \textit{TimeNet-48-Eps} performs better than existing benchmarks, while still being \textit{practically more feasible} as it looks at only up to 48 hours of current episode of a patient rather than the entire current episode.
%We also tried training an RNN encoder taking with all features providing univariate time series. Features from such an RNN encoder performed poorly compared to the TimeNet possibly due to 
For in-hospital mortality task, we observe comparable performance to existing benchmarks.

Training linear models is significantly fast and it took around 30 minutes for obtaining any of the binary classifiers while tuning for $\alpha \in [10^{-5}-10^{-3}]$ (five equally-spaced values) on a 32GB RAM machine with Quad Core i7 2.7GHz processor.

We observe that LASSO leads to  $96.2\pm0.8$ \% sparsity (i.e. percentage of weights $w_{jk}\approx 0$) for all classifiers leading to around 550 useful features (out of 13,680) for each phenotype classification.

\subsubsection{Relevance Scores for Raw Input Features}

%Sparse weights are easy to interpret and can give interesting insights into relevant features for a classification task (e.g. as used in \cite{micenkova2013explaining}).
%We map the TimeNet features with non-zero weights back to raw features: 
%We obtain relevance score of a raw input feature as the summation of the absolute weights given by the classifier to the 180 TimeNet-features corresponding to (the time series of) that input feature. 
We observe intuitive interpretation for relevance of raw input features using the weights assigned to various TimeNet features (refer Equation \ref{eq:relevance}): 
For example, as shown in Figure \ref{fig:relevance}, we obtain highest relevance scores for Glucose Level (feature 1) and Systolic Blood Pressure (feature 20) for Diabetes Mellitus with Complications (Figure \ref{fig:p1}), and  Essential Hypertension (Figure \ref{fig:p2}), respectively. 
Refer Supplementary Material Figure \ref{fig:feat_rel} for more details.
We conclude that \textit{even though TimeNet was never trained on MIMIC-III data, it still provides meaningful general-purpose features from time series of raw input features, and LASSO helps to select the most relevant ones for end-task by using labeled data.}
Further, extracting features using a deep recurrent neural network model for time series of each raw input feature independently -- rather than considering a multivariate time series -- eventually allows to easily assign relevance scores to raw features in the input domain, allowing a high-level basic model validation by domain-experts.

\section{Experimental Evaluation of HealthNet based Transfer Learning\label{ssec:healthnet}}
\subsection{Experimental Setup}
We evaluate HealthNet based Transfer Learning approach on same tasks as Section \ref{ssec:timenet} but in different setup.
%We evaluate the proposed approach on binary classification tasks as defined in \cite{harutyunyan2017multitask}: i) estimating the presence (class 1) or absence (class 0) of a phenotype (e.g. cardiac dysrhythmia, chronic kidney disease, etc.) from time series of parameters such as heart rate and respiratory rate, and ii) in-hospital mortality prediction where the goal is to predict whether the patient will survive or not given time series observations after ICU admission (class 1: patient dies, class 0: patient survives). 
%\newline\newline
%\textbf{Dataset details}\newline
%We use MIMIC-III (v1.4) clinical database \cite{johnson2016mimic} which consists of over 60,000 ICU stays across 40,000 critical care patients. 
%We use benchmark data from \cite{harutyunyan2017multitask} with same data-splits for train, validation and test datasets\footnote{Refer \cite{harutyunyan2017multitask} and https://github.com/yerevann/mimic3-benchmarks for dataset sizes and other details.}.
Train, validation and test sets for various scenarios considered %in our experiments 
are subsets of the respective original datasets (as described later).
%The benchmark dataset contains label information for presence/absence of 25 phenotypes common in adult ICUs (e.g. acute cerebrovascular disease, diabetes mellitus with complications, gastrointestinal hemorrhage, etc.). 
Out of 25, we consider $K=20$ phenotypes to obtain the pre-trained RNN which we refer to as HealthNet (HN), and test the transferability of the features from HN to remaining 5 phenotype (binary) classification tasks with varying labeled data sizes.
Since more than one phenotypes may be present in a patient at a time, we remove all patients with any of the 5 test phenotypes from the original train and validate sets (despite of them having one of the 20 train phenotypes also) to avoid any information leakage.
We report average results in terms of weighted AUROC (as in \cite{harutyunyan2017multitask}) on two random splits of 20 train phenotypes and 5 test phenotypes, such that we have 10 test phenotypes (tested one-at-a-time).
We also test transferability of HN features to in-hospital mortality prediction task.

We consider number of hidden layers $L=2$, batch size of 128, regularization using dropout factor \cite{pham2014dropout} of 0.3, and Adam optimizer \cite{kingma2014adam} with initial learning rate $10^{-4}$ for training RNNs. 
The number of hidden units $h$ with minimum $\mathcal{L}$ (eq. \ref{eq:RNN-C}) on the validation set is chosen from $\{100,200,300,400\}$. 
Best HN model was obtained for $h=300$ such that total number of features is $m=600$.
For fine-tuning of HN, we use same parameters as used in training HN and regularizer parameter as 0.01.
Linear classification model's L$_1$ parameter $\lambda$ is tuned on $\{0.1,1.0$,$\ldots$,$10^4\}$ (on a logarithmic scale) to minimize $\mathcal{L'}$ (eq. \ref{eq:RNN-C-tune}) on the validation set. 

\begin{figure}[ht]
	\centering
	\subfigure[\label{fig:phen_com}Phenotyping]{\includegraphics[width=0.5\columnwidth,trim={0cm 0cm 0cm 0cm},clip]{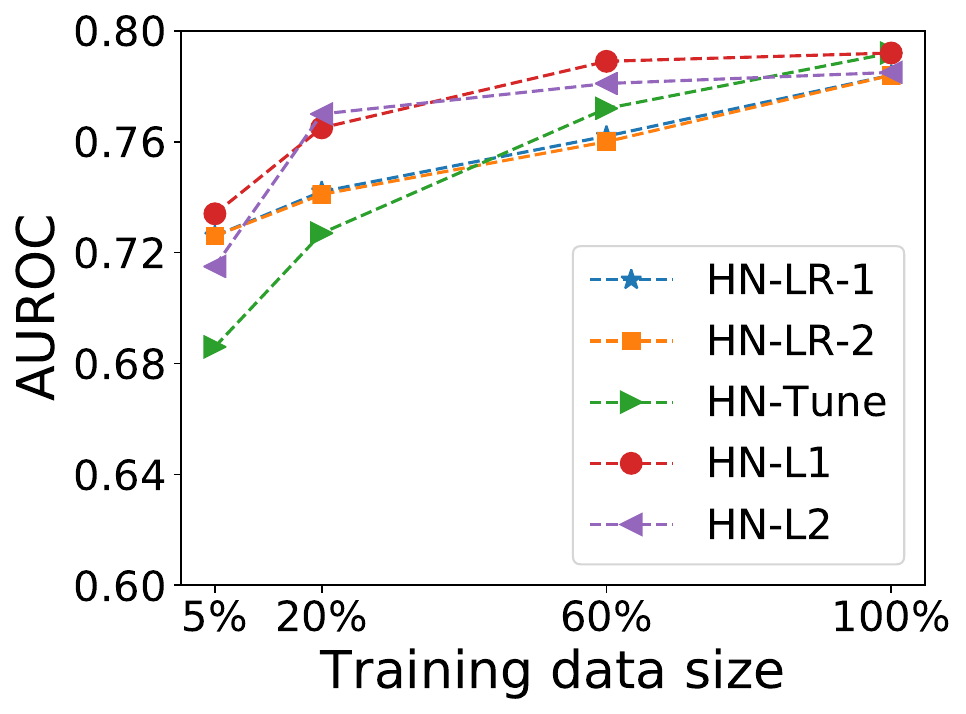}}~~
	\subfigure[\label{fig:mor_com}In-hospital mortality prediction]{\includegraphics[width=0.5\columnwidth,trim={0cm 0cm 0cm 0cm},clip]{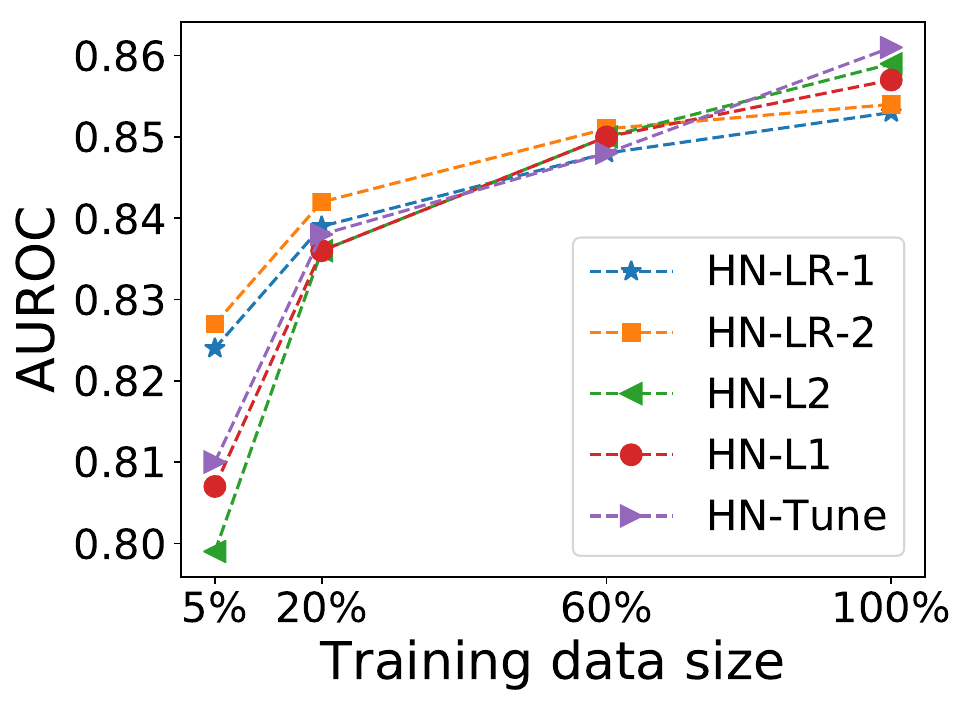}
	}
	\caption{Classification performance (AUROC with varying labeled data size) of HealthNet based transfer learning models. (a) HR-L1 outperforms for phenotyping task, (b) HN-LR-2 outperforms for in-hospital mortality task.}
\end{figure}

\begin{figure}[ht]
	\centering
	\subfigure[\label{fig:phen}Phenotyping]{\includegraphics[width=0.5\columnwidth,trim={0cm 0cm 0cm 0cm},clip]{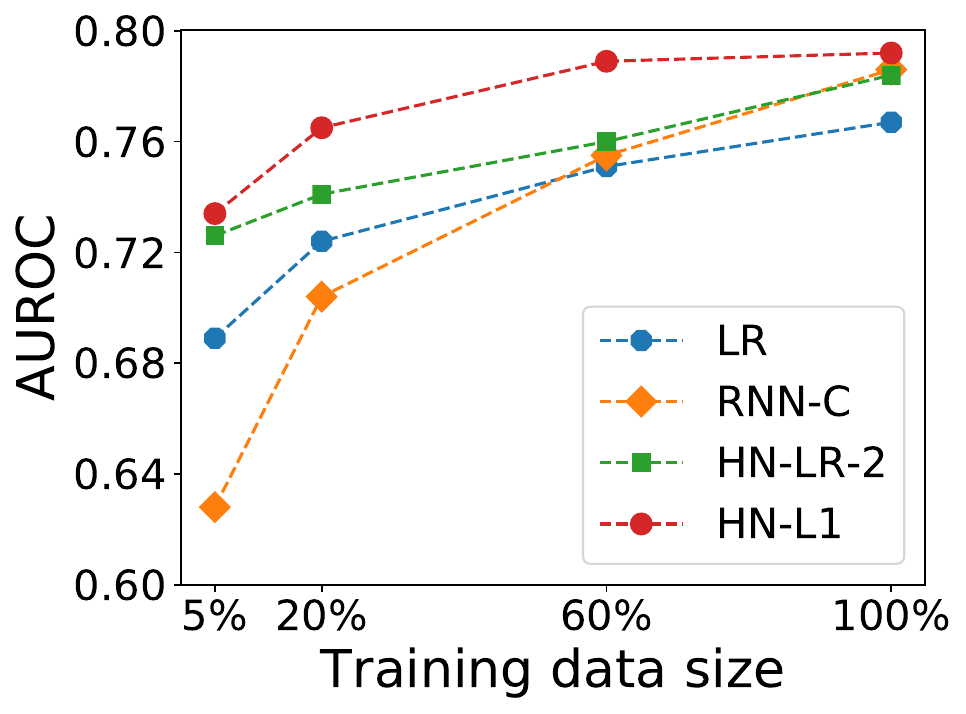}}~~
	\subfigure[\label{fig:mor}In-hospital mortality prediction]{\includegraphics[width=0.5\columnwidth,trim={0cm 0cm 0cm 0cm},clip]{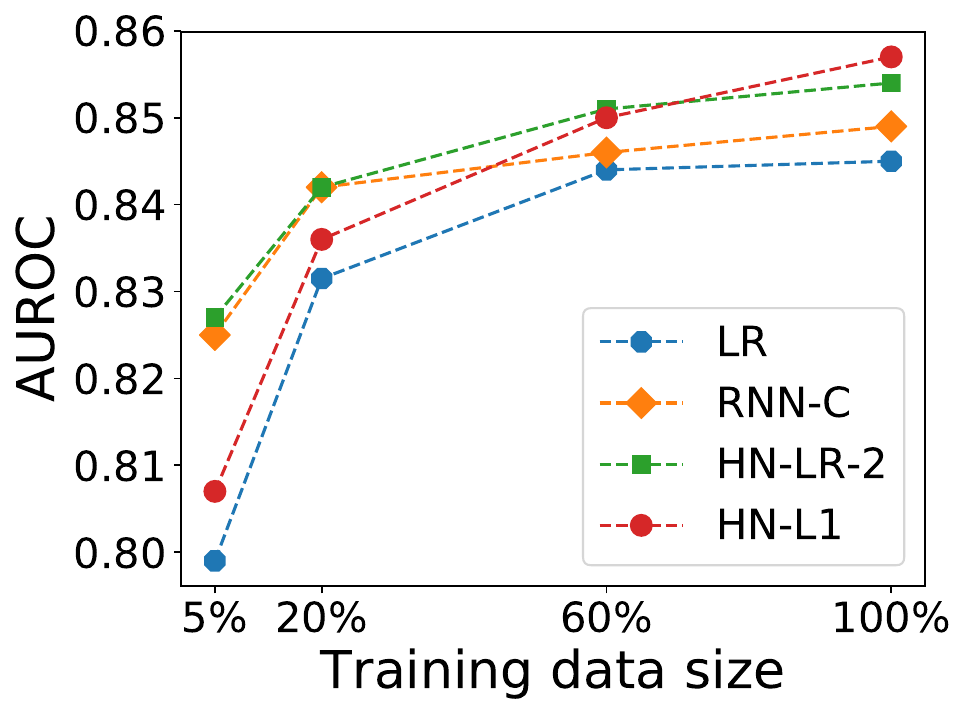}
	}
	\caption{Classification performance comparison: HealthNet based transfer learning models v/s baseline models (i.e. LR and RNN-C).}
\end{figure}

\subsection{Results and Observations}
We refer to the fine-tuned model using $L_1$ regularizer as \textbf{HN-L1}, $L_2$ regularizer as \textbf{HN-L2}, without regularizer as \textbf{HN-Tune}, and LR model learned using HN features as \textbf{HN-LR}, 
and consider two baselines for comparison: 1) Logistic Regression (\textbf{LR}) using statistical features (including mean, standard deviation, etc.) from raw time series as used in \cite{harutyunyan2017multitask}, 
2) RNN classifier (\textbf{RNN-C}) learned using training data for the target task.
To test the robustness of the models for small labeled training sets, we consider subsets of training and validation datasets, while the test set remains the same.
Further, we also evaluate the relevance of layer-wise features $\mathbf{z}_{\tau,l}$ from the $L=2$ hidden layers of HealthNet.
\textbf{HN-LR-1} and \textbf{HN-LR-2} refer to models trained using $\mathbf{z}_{\tau,2}$ (the topmost hidden layer only) and $\mathbf{z}_{\tau} = [\mathbf{z}_{\tau,1},\mathbf{z}_{\tau,2}]$ (from both hidden layers), respectively.
\begin{table}[t]
	\footnotesize
	\centering
	\footnotesize
	\caption{Fraction of features with weight $\approx$ 0. \label{tab:sparsity}}
	\begin{tabular}{|c|c|c|c|}
		\hline
		{\bfseries Task}&{\bfseries LR}&{\bfseries HN-LR-1}&{\bfseries HN-LR-2}\\
		\hline
		Phenotyping\footnote{The average and standard deviation over 10 phenotypes is reported.}&0.902 $\pm$ 0.023&0.955 $\pm$ 0.020&0.974 $\pm$ 0.011 \\
		\hline
		In-hospital mortality&0.917&0.787&0.867\\
		\hline
	\end{tabular}
%\footnotetxt{The average and standard deviation over 10 phenotypes is reported.}
	
\end{table}

\textbf{Comparison of HealthNet based transfer learning techniques}:
Phenotyping results in Figure \ref{fig:phen_com} suggests that (i) all our proposed transfer learning approaches perform equally well when using 100\% training data.
(ii) regularized fine-tuned models (HN-L1 and HN-L2) consistently outperform non-regularized fine-tuned model (HN-Tune) as training dataset is reduced. As the size of labeled training set reduced, non-regularized fine-tuned model is prone to overfitting 
due to large number of trainable parameters. %which degrade the performance in greater extent as compare to HN-L1 and HN-L2.
(iii) HN-L1 and HN-L2 outperform HN-LR model.

From Figure \ref{fig:mor_com}, we observe that (i) regularized fine-tuned models (HN-Tune, HN-L1, and HN-L2) perform comparable to HN-LR when training data is 100\%.
(ii) HN-LR consistently outperforms HN-Tune, HN-L1, and HN-L2 models as training dataset is reduced. 
This can be explained by the fact that the number of trainable parameters in HN-L1 and HN-L2 (of the order of square the number of hidden units) is significantly higher than the number of trianable parameters in HN-LR (of the order of number of hidden units), which is resulting in overfitting when labeled training dataset is very small.
%As even after applying regularizer, RNN model has more trainable parameters as compare to linear model which lead to overfitting on smaller training dataset.  

\textbf{Robustness to training data size}:
Phenotyping results in Figure \ref{fig:phen} suggest that: (i) HN-L1 and RNN-C perform equally well when using 100\% training data, and are better than LR.
This implies that \textit{the transfer learning based models are as effective as models trained specifically for the target task on large labeled datasets}. 
(ii) HN-L1 consistently outperforms RNN-C and LR models as training dataset is reduced.
As the size of labeled training dataset reduces, the performance of RNN-C as well as HN-L1 degrades. However, importantly, we observe that HN-LR degrades more gracefully and performs better than RNN-C. 
\textit{The performance gains from transfer learning are greater when the training set of the target task is small.
	Therefore, with transfer learning, fewer labeled instances are needed to achieve the same level of performance as model trained on target data alone.}
(iii) As labeled training set is reduced, LR performs better than RNN-C confirming that deep networks are prone to overfitting on small datasets.

From Figure \ref{fig:mor}, we interestingly observe that HN-LR-2 results are at least as good as RNN-C and LR on the seemingly unrelated task of mortality prediction, suggesting that \textit{the features learned are generic enough and transfer well.}

\textbf{Importance of features from different hidden layers}: We observe that HN-LR-1 and HN-LR-2 perform equally well for phenotyping task (Figure \ref{fig:phen_com}), suggesting that adding features $\mathbf{z}_{\tau,1}$ from lower hidden layer do not improve the performance given higher layer features $\mathbf{z}_{\tau,2}$. 
For the mortality prediction task, we observe slight improvement in HN-LR-2 over HN-LR-1, i.e. adding lower layer features helps. 
A possible explanation for this behavior is as follows: since training was done on phenotyping tasks, features from top-most layer suffice for new phenotypes as well; on the other hand, the more generic features from the lower layer are useful for the unrelated task of mortality prediction.

\textbf{Number of relevant features for a task}: We observe that only a small number of features are actually relevant for a target classification task out of large number of input features to LR models (714 for LR, 300 for HN-LR-1, and 600 for HN-LR-2), 
As shown in Table \ref{tab:sparsity}, $>$95\% of features have weight $\approx$ 0 (absolute value $<$ 0.001) for HN-LR models corresponding to phenotyping tasks due to sparsity constraint (eq. \ref{eq:RNN-C-tune}), i.e. most features do not contribute to the classification decision.
The weights of features that are non-zero for at least one of target tasks for HN-LR-1 are shown in Supplementary Material Figure \ref{fig:sparsity}.
We observe that, for example, for HN-LR-1 model only 130 features (out of 300) are relevant across the 10 phenotype classification tasks and the mortality prediction task. 
\textit{This suggests that HN provides several generic features while LR learns to select the most relevant ones given a small labeled dataset.}
Table \ref{tab:sparsity} and Figure \ref{fig:sparsity} also suggest that HN-LR models use larger number of features for mortality prediction task, possibly because concise features for mortality prediction are not available in the learned set of features as HN was pre-trained for phenotype identification tasks.

\section{Conclusion}\label{ssec:conclusion}
Deep neural networks require heavy computational resources for training and are prone to overfitting. Scarce labeled training data, significant hyper-parameter tuning efforts, and scarce computational resources are often a bottleneck in adopting deep learning based solutions to healthcare applications. % as 
In this work, we have proposed effective approaches for transfer learning in healthcare domain by using deep recurrent neural networks (RNN). 
We considered two scenarios for transfer learning: i) adapting a deep RNN-based universal time series feature extractor (TimeNet) to healthcare tasks and applications, and ii) adapting a deep RNN (HealthNet) pre-trained on healthcare tasks to a new related task.
Our approach brings the advantage of deep learning such as automated feature extraction and ability to easily deal with variable length time series while still being simple to adapt to the target tasks. 
We have demonstrated that our transfer learning approaches can lead to significant gains in classification performance compared to traditional models using carefully designed statistical features or task-specific deep models in scarcely labeled training data scenarios. 
Further, leveraging pre-trained models ensures very little tuning effort, and therefore, fast adaptation.
We also found that raw feature-wise handling of time series via TimeNet, and subsequent linear classifier training can provide insights into the importance and relevance of a raw feature (physiological parameter) for a given task while still modeling the temporal aspect. 
This raw feature relevance scoring can help domain-experts gain at least a high-level insight into the working of otherwise opaque deep RNNs.

In future, evaluating a domain-specific TimeNet-like model for clinical time series (e.g. trained only on MIMIC-III database) will be interesting.
Also, transferability and generalization capability of RNNs trained simultaneously on diverse tasks (such as length of stay, mortality prediction, phenotyping, etc. \cite{harutyunyan2017multitask,song2017attend}) to new tasks is an interesting future direction.

\iffalse
In this work, we leverage deep learning models efficiently via TimeNet for phenotyping and mortality prediction tasks, with little hyper-parameter tuning effort.
TimeNet-based features can be efficiently transferred to train linear interpretable classifiers for the end tasks considered while still achieving classification performance similar to more compute-intensive deep models trained from scratch. 

We have proposed an approach to leverage deep RNNs for small labeled datasets via transfer learning. 
We trained an RNN model to identify several phenotypes via multi-label classification. 
This model is found to be generalize well for new tasks including identification of new phenotypes, and interestingly, for mortality prediction.
We found that transfer learning performs better than the models trained specifically for the end task. 
Such transfer learning approaches can be a good starting point when building models with limited labeled datasets.
\fi

%\begin{acknowledgements}
%\end{acknowledgements}

\section*{Conflict of interest}
On behalf of all authors, the corresponding author states that there is no conflict of interest. 
\bibliographystyle{spbasic}      % basic style, author-year citations
\bibliography{phm-kdd2017,nips2016,online-ad,ijcai2017,kdd2018,isense,kdhd18,mlmh18,jhir2018}   % name your BibTeX data base

\begin{thebibliography}{33}
\providecommand{\natexlab}[1]{#1}
\providecommand{\url}[1]{{#1}}
\providecommand{\urlprefix}{URL }
\expandafter\ifx\csname urlstyle\endcsname\relax
  \providecommand{\doi}[1]{DOI~\discretionary{}{}{}#1}\else
  \providecommand{\doi}{DOI~\discretionary{}{}{}\begingroup
  \urlstyle{rm}\Url}\fi
\providecommand{\eprint}[2][]{\url{#2}}

\bibitem[{Bahdanau et~al.(2014)Bahdanau, Cho, and Bengio}]{bahdanau2014neural}
Bahdanau D, Cho K, Bengio Y (2014) {Neural machine translation by jointly
  learning to align and translate}. arXiv preprint arXiv:14090473

\bibitem[{Bengio(2012)}]{bengio2012deep}
Bengio Y (2012) Deep learning of representations for unsupervised and transfer
  learning. In: Proceedings of ICML Workshop on Unsupervised and Transfer
  Learning, pp 17--36

\bibitem[{Che et~al.(2016)Che, Purushotham, Cho, Sontag, and
  Liu}]{che2016recurrent}
Che Z, Purushotham S, Cho K, Sontag D, Liu Y (2016) {Recurrent neural networks
  for multivariate time series with missing values}. arXiv preprint
  arXiv:160601865

\bibitem[{Chen et~al.(2015)Chen, Keogh, Hu, Begum et~al.}]{UCRArchive}
Chen Y, Keogh E, Hu B, Begum N, et~al. (2015) The ucr time series
  classification archive. \url{www.cs.ucr.edu/~eamonn/time_series_data/}

\bibitem[{Cho et~al.(2014)Cho, Van~Merri{\"e}nboer, Gulcehre, Bahdanau,
  Bougares, Schwenk, and Bengio}]{cho2014learning}
Cho K, Van~Merri{\"e}nboer B, Gulcehre C, Bahdanau D, Bougares F, Schwenk H,
  Bengio Y (2014) {Learning phrase representations using RNN encoder-decoder
  for statistical machine translation}. arXiv preprint arXiv:14061078

\bibitem[{Choi et~al.(2016)Choi, Bahadori, Schuetz, Stewart, and
  Sun}]{choi2016doctor}
Choi E, Bahadori MT, Schuetz A, Stewart WF, Sun J (2016) Doctor ai: Predicting
  clinical events via recurrent neural networks. In: Machine Learning for
  Healthcare Conference, pp 301--318

\bibitem[{Gupta et~al.(2018{\natexlab{a}})Gupta, Malhotra, Vig, and
  Shroff}]{gupta2018using2}
Gupta P, Malhotra P, Vig L, Shroff G (2018{\natexlab{a}}) Transfer learning for
  clinical time series analysis using recurrent neural networks

\bibitem[{Gupta et~al.(2018{\natexlab{b}})Gupta, Malhotra, Vig, and
  Shroff}]{gupta2018using}
Gupta P, Malhotra P, Vig L, Shroff G (2018{\natexlab{b}}) Using features from
  pre-trained timenet for clinical predictions

\bibitem[{Harutyunyan et~al.(2017)Harutyunyan, Khachatrian, Kale, and
  Galstyan}]{harutyunyan2017multitask}
Harutyunyan H, Khachatrian H, Kale DC, Galstyan A (2017) Multitask learning and
  benchmarking with clinical time series data. arXiv preprint arXiv:170307771

\bibitem[{Hermans and Schrauwen(2013)}]{hermans2013training}
Hermans M, Schrauwen B (2013) Training and analysing deep recurrent neural
  networks. In: Advances in Neural Information Processing Systems, pp 190--198

\bibitem[{Hosmer~Jr et~al.(2013)Hosmer~Jr, Lemeshow, and
  Sturdivant}]{hosmer2013applied}
Hosmer~Jr DW, Lemeshow S, Sturdivant RX (2013) Applied logistic regression, vol
  398. John Wiley \& Sons

\bibitem[{Johnson et~al.(2016)Johnson, Pollard et~al.}]{johnson2016mimic}
Johnson AE, Pollard TJ, et~al. (2016) Mimic-iii, a freely accessible critical
  care database. Scientific data 3:160035

\bibitem[{Kingma and Ba(2014)}]{kingma2014adam}
Kingma DP, Ba J (2014) Adam: A method for stochastic optimization. arXiv
  preprint arXiv:14126980

\bibitem[{Lee et~al.(2017)Lee, Dernoncourt, and Szolovits}]{lee2017transfer}
Lee JY, Dernoncourt F, Szolovits P (2017) Transfer learning for named-entity
  recognition with neural networks. arXiv preprint arXiv:170506273

\bibitem[{Lipton et~al.(2015)Lipton, Kale, Elkan, and
  Wetzel}]{lipton2015learning}
Lipton ZC, Kale DC, Elkan C, Wetzel R (2015) Learning to diagnose with lstm
  recurrent neural networks. arXiv preprint arXiv:151103677

\bibitem[{Malhotra et~al.(2015)Malhotra, Vig, Shroff, and Agarwal}]{p:lstm-ad}
Malhotra P, Vig L, Shroff G, Agarwal P (2015) {Long Short Term Memory Networks
  for Anomaly Detection in Time Series}. In: {ESANN, 23rd European Symposium on
  Artificial Neural Networks, Computational Intelligence and Machine Learning},
  pp 89--94

\bibitem[{Malhotra et~al.(2017)Malhotra, TV, Vig, Agarwal, and
  Shroff}]{malhotra2017timenet}
Malhotra P, TV V, Vig L, Agarwal P, Shroff G (2017) {TimeNet: Pre-trained deep
  recurrent neural network for time series classification}. In: {25th European
  Symposium on Artificial Neural Networks, Computational Intelligence and
  Machine Learning}, pp {607--612}

\bibitem[{Micenkov{\'a} et~al.(2013)Micenkov{\'a}, Dang, Assent, and
  Ng}]{micenkova2013explaining}
Micenkov{\'a} B, Dang XH, Assent I, Ng RT (2013) Explaining outliers by
  subspace separability. In: Data Mining (ICDM), 2013 IEEE 13th International
  Conference on, IEEE, pp 518--527

\bibitem[{Miotto et~al.(2016)Miotto, Li, Kidd, and Dudley}]{miotto2016deep}
Miotto R, Li L, Kidd BA, Dudley JT (2016) Deep patient: an unsupervised
  representation to predict the future of patients from the electronic health
  records. Scientific reports 6:26094

\bibitem[{Miotto et~al.(2017)Miotto, Wang, Wang, Jiang, and
  Dudley}]{miotto2017deep}
Miotto R, Wang F, Wang S, Jiang X, Dudley JT (2017) Deep learning for
  healthcare: review, opportunities and challenges. Briefings in bioinformatics

\bibitem[{Nguyen et~al.(2017)Nguyen, Tran, Wickramasinghe, and
  Venkatesh}]{nguyen2017mathtt}
Nguyen P, Tran T, Wickramasinghe N, Venkatesh S (2017) Deepr: A convolutional
  net for medical records. IEEE journal of biomedical and health informatics
  21(1):22--30

\bibitem[{Pan and Yang(2010)}]{pan2010survey}
Pan SJ, Yang Q (2010) A survey on transfer learning. IEEE Transactions on
  knowledge and data engineering 22(10):1345--1359

\bibitem[{Pascanu et~al.(2013)Pascanu, Mikolov, and
  Bengio}]{bengio2013regularizer}
Pascanu R, Mikolov T, Bengio Y (2013) On the difficulty of training recurrent
  neural networks. arXiv preprint arXiv:12115063

\bibitem[{Pham et~al.(2014)Pham, Bluche, Kermorvant, and
  Louradour}]{pham2014dropout}
Pham V, Bluche T, Kermorvant C, Louradour J (2014) Dropout improves recurrent
  neural networks for handwriting recognition. In: {Frontiers in Handwriting
  Recognition (ICFHR)}, IEEE, pp 285--290

\bibitem[{Purushotham et~al.(2017)Purushotham, Meng, Che, and
  Liu}]{purushotham2017benchmark}
Purushotham S, Meng C, Che Z, Liu Y (2017) Benchmark of deep learning models on
  large healthcare mimic datasets. arXiv preprint arXiv:171008531

\bibitem[{Rajkomar et~al.(2018)Rajkomar, Oren, Chen, Dai, Hajaj, Liu, Liu, Sun,
  Sundberg, Yee et~al.}]{rajkomar2018scalable}
Rajkomar A, Oren E, Chen K, Dai AM, Hajaj N, Liu PJ, Liu X, Sun M, Sundberg P,
  Yee H, et~al. (2018) Scalable and accurate deep learning for electronic
  health records. arXiv preprint arXiv:180107860

\bibitem[{Rav{\`\i} et~al.(2017)Rav{\`\i}, Wong, Deligianni, Berthelot,
  Andreu-Perez, Lo, and Yang}]{ravi2017deep}
Rav{\`\i} D, Wong C, Deligianni F, Berthelot M, Andreu-Perez J, Lo B, Yang GZ
  (2017) Deep learning for health informatics. IEEE journal of biomedical and
  health informatics 21(1):4--21

\bibitem[{Serra et~al.(2018)Serra, Pascual, and Karatzoglou}]{serra2018}
Serra J, Pascual S, Karatzoglou A (2018) Towards a universal neural network
  encoder for time series

\bibitem[{Simonyan and Zisserman(2014)}]{simonyan2014very}
Simonyan K, Zisserman A (2014) Very deep convolutional networks for large-scale
  image recognition. arXiv preprint arXiv:14091556

\bibitem[{Song et~al.(2017)Song, Rajan, Thiagarajan, and
  Spanias}]{song2017attend}
Song H, Rajan D, Thiagarajan JJ, Spanias A (2017) Attend and diagnose: Clinical
  time series analysis using attention models. arXiv preprint arXiv:171103905

\bibitem[{Sutskever et~al.(2014)Sutskever, Vinyals, and
  Le}]{sutskever2014sequence}
Sutskever I, Vinyals O, Le QV (2014) Sequence to sequence learning with neural
  networks. In: {Advances in Neural Information Processing Systems}, pp
  3104--3112

\bibitem[{Tibshirani(1996)}]{tibshirani1996regression}
Tibshirani R (1996) Regression shrinkage and selection via the lasso. Journal
  of the Royal Statistical Society Series B (Methodological) pp 267--288

\bibitem[{Yosinski et~al.(2014)Yosinski, Clune, Bengio, and
  Lipson}]{yosinski2014transferable}
Yosinski J, Clune J, Bengio Y, Lipson H (2014) How transferable are features in
  deep neural networks? In: Advances in neural information processing systems,
  pp 3320--3328

\end{thebibliography}

\appendix
\section*{Supplementary Material}
\section{Multilayered RNN with Gated Recurrent Units}\label{apdx:gru_missing}

A Gated Recurrent Unit (GRU) \cite{cho2014learning} consists of an \textit{update gate} and a \textit{reset gate} that control the flow of information by manipulating the \textit{hidden state} of the unit as in Equation~\ref{eq:reset_gate}.

In an RNN with $L$ hidden layers, the reset gate is used to compute a proposed value $\mathbf{\tilde z}_t^l$ for the hidden state $\mathbf{z}_t^l$ at time $t$ for the $l$-th hidden layer by using the hidden state $\mathbf{z}_{t-1}^l$ and the hidden state $\mathbf{z}_{t}^{l-1}$ of the units in the lower hidden layer at time $t$.
The update gate decides as to what fractions of previous hidden state and proposed hidden state to use to obtain the updated hidden state at time $t$.
In turn, the values of the reset gate and update gate themselves depend on the $\mathbf{z}_{t-1}^l$ and $\mathbf{z}_{t}^{l-1}$

We use dropout variant for RNNs as proposed in \cite{pham2014dropout} for regularization such that dropout is applied only to the non-recurrent connections, ensuring information flow across time-steps. 

The time series goes through the following transformations iteratively for $t=1$ through $T$, where $T$ is length of the time series:
\begin{equation}\label{eq:reset_gate}
\begin{aligned}
reset\, gate: \mathbf{r}_t^l = \sigma(\mathbf{W}_r^l\cdot\mathbf{D}(\mathbf{z}_{t}^{l-1}),\mathbf{z}_{t-1}^l])\\
update\, gate: \mathbf{u}_t^l = \sigma(\mathbf{W}_u^l\cdot[\mathbf{D}(\mathbf{z}_{t}^{l-1}),\mathbf{z}_{t-1}^l])\\
proposed\, state: \mathbf{\tilde z}_t^l = \tanh(\mathbf{W}_p^l\cdot[\mathbf{D}(\mathbf{z}_{t}^{l-1}),\mathbf{r}_t\odot \mathbf{z}_{t-1}^l])\\
hidden\, state: \mathbf{z}_t^l = (1-\mathbf{u}_t^l)\odot \mathbf{z}_{t-1}^l+\mathbf{u}_t^l\odot\mathbf{\tilde z}_t^l\\
\end{aligned}
\end{equation}
where $\odot$ is Hadamard product, $[\mathbf{a},\mathbf{b}]$ is concatenation of vectors $\mathbf{a}$ and $\mathbf{b}$, $\mathbf{D}(\cdot)$ is dropout operator that randomly sets the dimensions of its argument to zero with probability equal to dropout rate, $\mathbf{z}_{t}^{0}$ equals the input at time $t$. $\mathbf{W}_r$, $\mathbf{W}_u$, and $\mathbf{W}_p$ are weight matrices of appropriate dimensions s.t. $\mathbf{r}_t^l, \mathbf{u}_t^l, \mathbf{\tilde z}_t^l$, and $\mathbf{z}_t^l$ are vectors in $\mathbf{R}^{c^l}$, where $c^l$ is the number of units in layer $l$. 
The sigmoid ($\sigma$) and $tanh$ activation functions are applied element-wise. 

%\section*{Phenotype-wise results}
\begin{figure*}[h]
	\includegraphics[width=\textwidth]{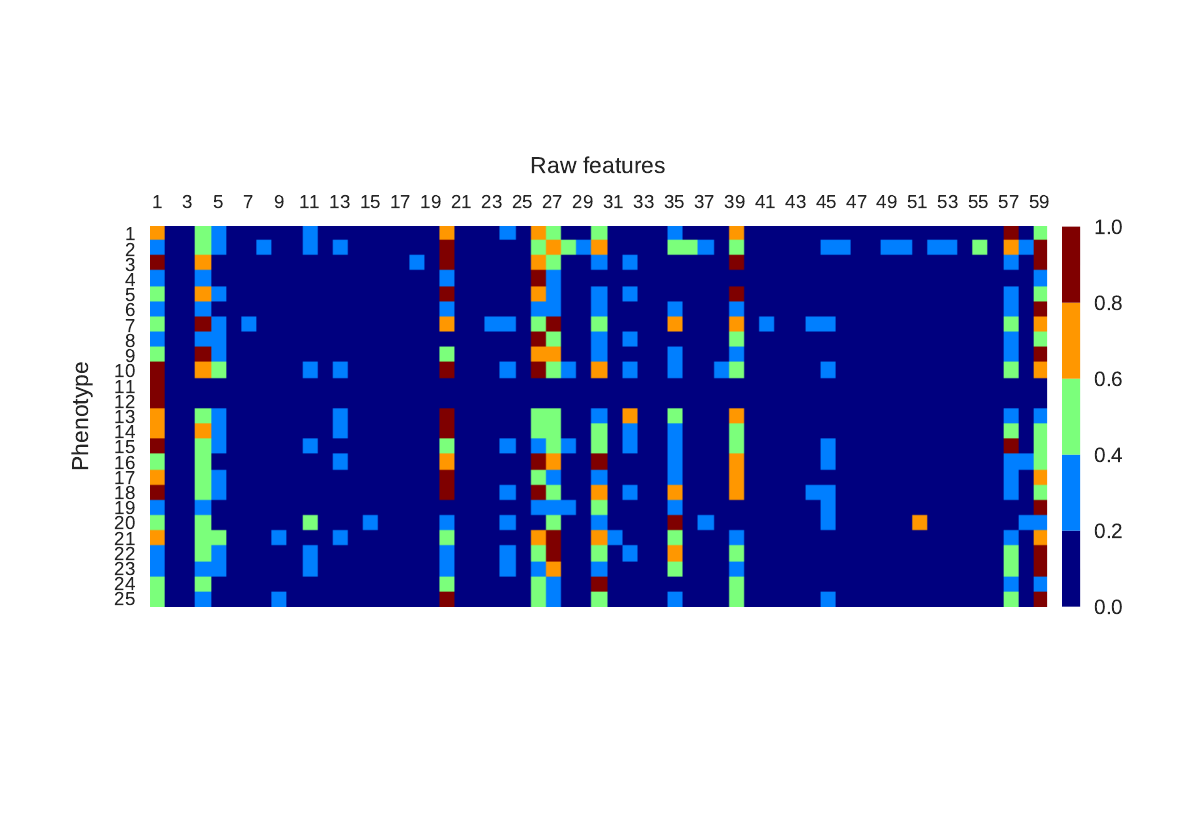}
	\vspace*{-25mm}
	\caption{\label{fig:rel} Feature relevance scores for 25 phenotypes using TimeNet based transfer learning. Refer Table \ref{tab:phenotype-wise} for names of phenotypes, and Table \ref{tab:feature} for names of raw features.\label{fig:feat_rel}}
\end{figure*}

\begin{figure*}[h]
	\centering
	\includegraphics[trim={0cm 0.5cm 0cm 0cm},clip,width=\textwidth]{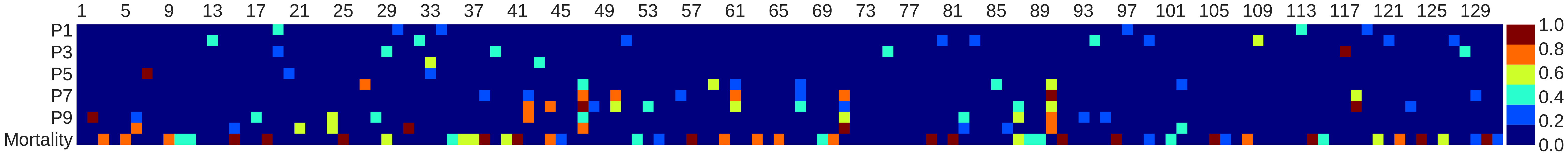}
	\caption{\label{fig:sparsity} Feature weights (absolute) for HN-LR-1. Here P$_i$ ($i=1,\ldots,10$) denotes $i$-th phenotype identification task. x-axis: Feature Number, y-axis: Task.}
\end{figure*}

\begin{table*}[ht]
	\centering
	\footnotesize
	\caption{Phenotype-wise Classification Performance in terms of AUROC.\label{tab:phenotype-wise}}
	\Rotatebox{90}{%
	\begin{tabular}{|c|c|p{0.1\linewidth}|c|p{0.1\linewidth}|p{0.1\linewidth}|p{0.1\linewidth}|}
		%\toprule
		\hline
		\textbf{S.No.}&\textbf{Phenotype}&{\bfseries LSTM-Multi}&{\bfseries TimeNet-48}&{\bfseries TimeNet-All}&{\bfseries TimeNet-48-Eps}&{\bfseries TimeNet-All-Eps}\\
		\hline
		1&Acute and unspecified renal failure&0.8035&0.7861&0.7887&0.7912&0.7941\\
		\hline
		2&Acute cerebrovascular disease&0.9089&0.8989&0.9031&0.8986&0.9033\\
		\hline
		3&Acute myocardial infarction&0.7695&0.7501&0.7478&0.7533&0.7509\\
		\hline
		4&Cardiac dysrhythmias&0.684&0.6853&0.7005&0.7096&0.7239\\
		\hline
		5&Chronic kidney disease&0.7771&0.7764&0.7888&0.7960&0.8061\\
		\hline
		6&Chronic obstructive pulmonary disease and bronchiectasis&0.6786&0.7096&0.7236&0.7460&0.7605\\
		\hline
		7&Complications of surgical procedures or medical care&0.7176&0.7061&0.6998&0.7092&0.7029\\
		\hline
		8&Conduction disorders&0.726&0.7070&0.7111&0.7286&0.7324\\
		\hline
		9&Congestive heart failure; nonhypertensive&0.7608&0.7464&0.7541&0.7747&0.7805\\
		\hline
		10&Coronary atherosclerosis and other heart disease&0.7922&0.7764&0.7760&0.8007&0.8016\\
		\hline
		11&Diabetes mellitus with complications&0.8738&0.8748&0.8800&0.8856&0.8887\\
		\hline
		12&Diabetes mellitus without complication&0.7897&0.7749&0.7853&0.7904&0.8000\\
		\hline
		13&Disorders of lipid metabolism&0.7213&0.7055&0.7119&0.7217&0.7280\\
		\hline
		14&Essential hypertension&0.6779&0.6591&0.6650&0.6757&0.6825\\
		\hline
		15&Fluid and electrolyte disorders&0.7405&0.7351&0.7301&0.7377&0.7328\\
		\hline
		16&Gastrointestinal hemorrhage&0.7413&0.7364&0.7309&0.7386&0.7343\\
		\hline
		17&Hypertension with complications and secondary hypertension&0.76&0.7606&0.7700&0.7792&0.7871\\
		\hline
		18&Other liver diseases&0.7659&0.7358&0.7332&0.7573&0.7530\\
		\hline
		19&Other lower respiratory disease&0.688&0.6847&0.6897&0.6896&0.6922\\
		\hline
		20&Other upper respiratory disease&0.7599&0.7515&0.7565&0.7595&0.7530\\
		\hline
		21&Pleurisy; pneumothorax; pulmonary collapse&0.7027&0.6900&0.6882&0.6909&0.6997\\
		\hline
		22&Pneumonia &0.8082&0.7857&0.7916&0.7890&0.7943\\
		\hline
		23&Respiratory failure; insufficiency; arrest (adult)&0.9015&0.8815&0.8856&0.8834&0.8876\\
		\hline
		24&Septicemia (except in labor)&0.8426&0.8276&0.8140&0.8296&0.8165\\
		\hline
		25&Shock&0.876&0.8764&0.8564&0.8763&0.8562\\
		\hline
	\end{tabular}
}%
\end{table*}

\begin{table*}[h]
	\centering
	\footnotesize
	\caption{List of raw input features\label{tab:feature}.}
	\Rotatebox{90}{%
	\begin{tabular}{|c|c|c|c|}
		\hline 
		%\textbf{S.No.}&\textbf{Raw Feature (as used in \cite{harutyunyan2017multitask})}&&\\
		%\hline
		1&Glucose&31&Glascow coma scale eye opening $\rightarrow$ 3 To speech\\
		2&Glascow coma scale total $\rightarrow$ 7&32&Height\\
		3&Glascow coma scale verbal response $\rightarrow$ Incomprehensible sounds&33&Glascow coma scale motor response $\rightarrow$ 5 Localizes Pain\\
		4&Diastolic blood pressure&34&Glascow coma scale total $\rightarrow$ 14\\
		5&Weight&35&Fraction inspired oxygen\\
		6&Glascow coma scale total $\rightarrow$ 8&36&Glascow coma scale total $\rightarrow$ 12\\
		7&Glascow coma scale motor response $\rightarrow$ Obeys Commands&37&Glascow coma scale verbal response $\rightarrow$ Confused\\
		8&Glascow coma scale eye opening $\rightarrow$ None&38&Glascow coma scale motor response $\rightarrow$ 1 No Response\\
		
		9&Glascow coma scale eye opening $\rightarrow$ To Pain&39&Mean blood pressure\\
		10&Glascow coma scale total $\rightarrow$ 6&40&Glascow coma scale total $\rightarrow$ 4\\
		
		11&Glascow coma scale verbal response $\rightarrow$ 1.0 ET/Trach&41&Glascow coma scale eye opening $\rightarrow$ To Speech\\
		12& Glascow coma scale total $\rightarrow$ 5&42&Glascow coma scale total $\rightarrow$ 15\\
		13&Glascow coma scale verbal response $\rightarrow$ 5 Oriented&43&Glascow coma scale motor response $\rightarrow$ 4 Flex-withdraws\\
		14&Glascow coma scale total $\rightarrow$ 3&44&Glascow coma scale motor response $\rightarrow$ No response\\
		15&Glascow coma scale verbal response $\rightarrow$ No Response&45&Glascow coma scale eye opening $\rightarrow$ Spontaneously\\
		16&Glascow coma scale motor response $\rightarrow$ 3 Abnorm flexion&46&Glascow coma scale verbal response $\rightarrow$ 4 Confused\\
		17&Glascow coma scale verbal response $\rightarrow$ 3 Inapprop words&47&Capillary refill rate $\rightarrow$ 0.0\\
		18&Capillary refill rate $\rightarrow$ 1.0&48&Glascow coma scale total $\rightarrow$ 13\\
		19&Glascow coma scale verbal response $\rightarrow$ Inappropriate Words&49&Glascow coma scale eye opening $\rightarrow$ 1 No Response\\
		20&Systolic blood pressure&50&Glascow coma scale motor response $\rightarrow$ Abnormal extension\\
		21&Glascow coma scale motor response $\rightarrow$ Flex-withdraws&51&Glascow coma scale total $\rightarrow$ 11\\
		22&Glascow coma scale total $\rightarrow$ 10&52&Glascow coma scale verbal response $\rightarrow$ 2 Incomp sounds\\
		23&Glascow coma scale motor response $\rightarrow$ Obeys Commands&53&Glascow coma scale total $\rightarrow$ 9\\
		24&Glascow coma scale verbal response $\rightarrow$ No Response-ETT&54&Glascow coma scale motor response $\rightarrow$ Abnormal Flexion\\
		25&Glascow coma scale eye opening $\rightarrow$ 2 To pain&55&Glascow coma scale verbal response $\rightarrow$ 1 No Response\\
		26&Heart Rate&56&Glascow coma scale motor response $\rightarrow$ 2 Abnorm extensn\\
		27&Respiratory rate&57&pH\\
		28&Glascow coma scale verbal response $\rightarrow$ Oriented&58&Glascow coma scale eye opening $\rightarrow$ 4 Spontaneously\\
		29&Glascow coma scale motor response $\rightarrow$ Localizes Pain&59&Oxygen saturation\\
		30&Temperature&&\\

		\hline
	\end{tabular}
	}%
\end{table*}

\end{document}